\documentclass[lettersize,journal]{IEEEtran}

\usepackage[dvipsnames,table]{xcolor}
\usepackage{amsmath,amsfonts,amssymb}
\usepackage{bm}
\usepackage{graphicx}
\usepackage{booktabs}
\usepackage{multirow}
\usepackage{makecell}
\usepackage{pifont}
\usepackage{enumitem}
\usepackage{array}
\usepackage{textcomp}
\usepackage{url}
\usepackage{stfloats}
\usepackage{cite}
\usepackage{hyperref}
\usepackage{newtxtext}
\usepackage{lmodern}
\usepackage[ruled,vlined]{algorithm2e}

\newif\ifshowrevisions
\showrevisionstrue
\ifshowrevisions
  
\else
  
\fi

\colorlet{colorFst}{Green!25}       
\colorlet{colorSnd}{SpringGreen!45} 
\colorlet{colorTrd}{Yellow!30}      

\makeatletter
\newcommand{\MAGSteaser}{%
  \vspace{0.4\baselineskip}%
  \begingroup
  \centering
  \includegraphics[width=0.99\textwidth]{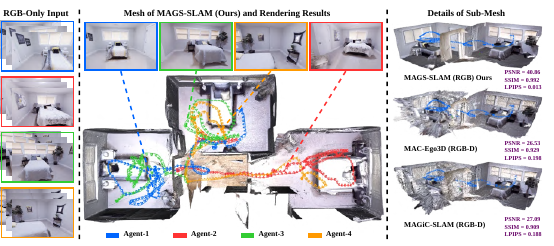}\par
  \vspace{-0.2\baselineskip}%
  \def\@captype{figure}%
  \setcounter{figure}{0}\refstepcounter{figure}
  \@makecaption{\figurename~\thefigure}{%
    \textbf{Overview of MAGS-SLAM.}
    From only the RGB video streams of multiple agents (\emph{left}), MAGS-SLAM reconstructs a single globally consistent photorealistic 3D Gaussian map without any active depth sensor or {inter-agent extrinsic calibration}.
    The \emph{middle} panel overlays the color-coded agent trajectories on the fused mesh together with per-agent renderings, showing that the independently scaled monocular submaps are aligned into one coherent scene.
    The \emph{right} panel zooms into a shared region: relying on RGB only, MAGS-SLAM surpasses the rendering quality of the state-of-the-art RGB-D systems MAC-Ego3D~\cite{xu2025mac} and MAGiC-SLAM~\cite{yugay2025magic}. The project website of MAGS-SLAM is available at {\textcolor{magenta}{{\href{https://mags-slam.github.io}{https://mags-slam.github.io}}}}. }\par
  \endgroup
  \vspace{-1.8\baselineskip}%
}
\g@addto@macro\@maketitle{\MAGSteaser}
\makeatother

\begin{document}

\title{MAGS-SLAM: Monocular Multi-Agent Gaussian Splatting SLAM for Geometrically and Photometrically Consistent Reconstruction}

\author{Zhihao Cao$^{1, \ast}$, Qi Shao$^{2, 3, \ast}$, Shuhao Zhai$^{4, \ast}$, Jing Zhang$^{5}$, Anh Nguyen$^{2}$, Hesheng Wang$^{6}$, Baoru Huang$^{2, 7, \dag}$%
\thanks{$^{\ast}$Equal contribution; $^{\dag}$Corresponding author.}%
\thanks{$^{1}$ETH Zurich, Switzerland; $^{2}$University of Liverpool, United Kingdom; $^{3}$Harbin Engineering University, China; $^{4}$University of Ottawa, Canada; $^{5}$Wuhan University, China; $^{6}$Shanghai Jiao Tong University, China; $^{7}$Imperial College, United Kingdom.}%
}

\markboth{IEEE Transactions on Automation Science and Engineering}%
{IEEE Transactions on Automation Science and Engineering}

\maketitle

\begin{abstract}
Collaborative photorealistic 3D reconstruction from multiple agents enables rapid large-scale scene capture for virtual production and cooperative multi-robot exploration. While recent 3D Gaussian Splatting (3DGS) SLAM algorithms can generate high-fidelity real-time mapping, most of the existing multi-agent Gaussian SLAM methods still rely on RGB-D sensors to obtain metric depth and simplify cross-agent alignment, limiting their deployment on low-cost or power-constrained robotic platforms, especially given the wider availability of RGB cameras.
To address this challenge, we propose MAGS-SLAM, the first RGB-only multi-agent 3DGS SLAM framework for collaborative scene reconstruction.
Each agent independently builds local monocular Gaussian submaps and transmits compact submap summaries rather than raw observations.
To facilitate robust collaboration in the presence of monocular scale ambiguity, our framework integrates compact submap communication, geometry- and appearance-aware loop verification, and occupancy-aware Gaussian fusion, enabling coherent global reconstruction without active depth sensors.
We further introduce ReplicaMultiagent Plus, a benchmark containing larger robot teams for evaluating collaborative Gaussian SLAM. {Extensive experiments on both synthetic and real-world datasets demonstrate that MAGS-SLAM, using RGB images alone, achieves tracking accuracy, rendering quality, and geometric quality that are competitive with or superior to those of state-of-the-art RGB-D collaborative Gaussian SLAM methods.}
\end{abstract}

\begingroup
\renewcommand{\abstractname}{Note to Practitioners}
\begin{abstract}
This work is motivated by the need for teams of robots to jointly build a photorealistic 3D map of a shared workspace using only ordinary RGB cameras. Existing collaborative dense-mapping systems depend on RGB-D depth sensors, which add cost, power consumption, and calibration burden to each platform. MAGS-SLAM addresses this challenge: each agent runs a monocular SLAM front-end and exchanges compact submap summaries, while a coordinator aligns the independently scaled maps and fuses them into a single consistent 3D Gaussian map that can be rendered in real time. Practitioners can deploy the system on camera-only platforms such as micro aerial vehicles, AR headsets, or low-cost ground robots for applications including cooperative inspection, warehouse digitization, and virtual production. 
{In a practical automated-mapping workflow, multiple camera-only robots concurrently explore a multi-room workspace. Without requiring depth sensors, each robot operates on a commodity GPU, performs real-time tracking, and transmits compact submap summaries, enabling efficient deployment over standard wireless networks.
However, }the current implementation assumes reliable connectivity to a coordinator and indoor-scale scenes; extending the approach to intermittent communication and very large environments remains future work.
\end{abstract}
\endgroup



\section{Introduction}
\label{sec:intro}

\IEEEPARstart{C}{ollaborative} photorealistic 3D reconstruction from multiple agents enables rapid large-scale scene capture, critical for virtual production, immersive telepresence, multi-user augmented reality, cooperative multi-robot exploration, {and automation applications such as automated as-built verification and infrastructure inspection \cite{hu2023cp, tian2022kimera, yugay2025magic, deng2025mne}.}
Recently, 3D Gaussian Splatting (3DGS)~\cite{kerbl20233d} has emerged as a powerful representation for real-time photorealistic scene reconstruction and dense visual SLAM.
As an explicit, differentiable, and real-time renderable scene representation, 3DGS supports efficient map optimization and high-fidelity view synthesis, motivating a rapidly growing family of single-agent Gaussian SLAM systems~\cite{keetha2024splatam, yan2024gs, matsuki2024gaussian, yugay2023gaussian, huang2024photo, ha2024rgbd, peng2024rtg, li2024sgs, hu2024cg, zhu2025loopsplat, sandstrom2025splat}.
However, a single agent observes only a limited portion of an environment at any time, rendering sequential capture inefficient for large-scale or multi-room scenes.
Multi-agent collaborative SLAM addresses this bottleneck by enabling parallel exploration, complementary viewpoints, and redundant observations.
Early collaborative SLAM systems mainly focused on sparse geometric maps~\cite{schmuck2019ccm, lajoie2020door, tian2022kimera, lajoie2023swarm}, while recent methods have extended collaboration to renderable neural implicit and Gaussian representations~\cite{hu2023cp, deng2025mne, deng2025mcn, yugay2025magic, xu2025mac, thomas2025grand, chen2026coma}.
Nevertheless, existing multi-agent 3DGS systems still require RGB-D sensors to obtain metric depth and simplify inter-agent alignment to rigid $\mathrm{SE}(3)$ registration.
This dependence on active depth sensing increases hardware cost, power consumption, calibration complexity, and synchronization overhead, especially given the wider availability of RGB cameras on most robotic platforms.
Consequently, monocular RGB-only multi-agent Gaussian SLAM remains a critical open challenge, as monocular agents reconstruct local Gaussian maps with inherent scale ambiguity and uncertain cross-agent overlaps.

To address this gap, we propose \textbf{MAGS-SLAM}, the first monocular RGB-only multi-agent 3DGS SLAM framework for collaborative photorealistic reconstruction. Each agent independently estimates monocular poses and depths, incrementally constructs local Gaussian submaps, and communicates compact submap summaries instead of raw observations or dense maps. To bridge independently scaled monocular maps, a communication-efficient coordinator establishes reliable cross-agent Sim(3) constraints from compact submap summaries through descriptor retrieval, spatially well-conditioned geometric verification, and sparse-cloud refinement. These constraints define a global Sim(3) submap graph that aligns monocular reconstructions with both geometric and photometric consistency. An asynchronous occupancy-aware fusion stage then fuses the aligned Gaussian submaps, filtering redundant Gaussians while preserving observed free space, followed by joint pose-Gaussian photometric refinement. The resulting system achieves competitive tracking accuracy and photorealistic reconstruction without any depth sensors.

Our main contributions are:
\begin{enumerate}
    [itemsep=2pt,topsep=2pt,leftmargin=15pt]
    \item We propose the first monocular RGB-only multi-agent 3D Gaussian Splatting SLAM system. It integrates Gaussian front-ends, compact submap summaries, inter-agent verification, $\mathrm{Sim}(3)$ submap pose graph, and occupancy-aware fusion into a unified framework, achieving accurate tracking and photorealistic reconstruction.

    \item We propose a Multi-agent Pose-Graph Bundle Adjustment (MaPGBA)-consistent $\mathrm{Sim}(3)$ loop closure mechanism for multi-agent systems, which jointly resolves intra- and inter-agent scale drift through a submap-level $\mathrm{Sim}(3)$ pose graph coupling geometric and photometric residuals.

    \item We propose an occupancy-aware fusion framework for coherent multi-agent Gaussian maps. It combines occupancy-grid deduplication, decoupled coordinator, and joint pose-Gaussian photometric refinement to {reduce} duplicated Gaussians, and photometric seams across agents.

    \item We introduce \textbf{ReplicaMultiagent Plus} dataset. While existing multi-agent datasets are typically limited to 2-3 agents with short trajectories, our dataset scales to 4 agents with long-horizon trajectories. In addition, we provide ground-truth geometry and semantic annotations, supporting the evaluation of monocular, RGB-D, and semantic multi-agent SLAM for collaborative dense reconstruction.
\end{enumerate}

\begin{figure*}[!t]
  \centering
  \includegraphics[width=\textwidth]{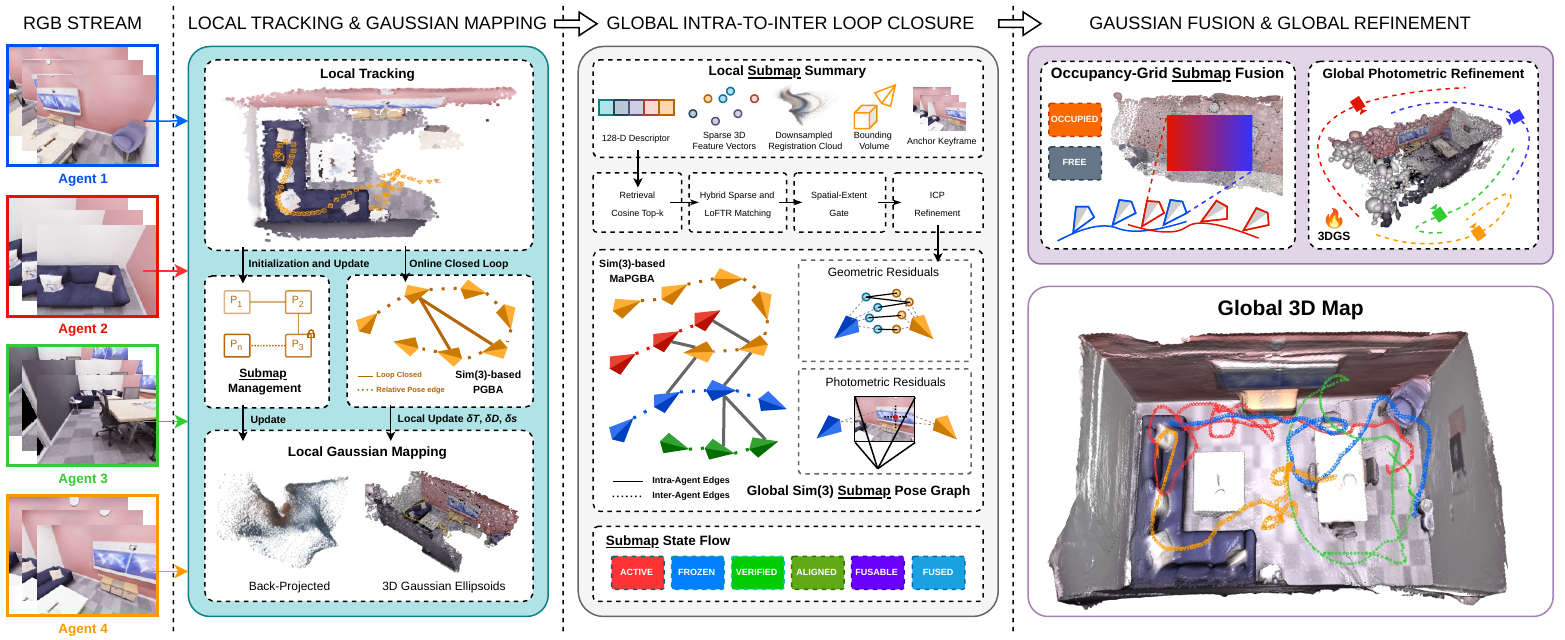}
  \vspace{-1.6\baselineskip}
  \caption{
    \textbf{MAGS-SLAM pipeline.} Given only the RGB stream of each agent (\emph{far left}), the system proceeds in three stages.
    \textbf{(1) Local tracking and Gaussian mapping} (per agent): Joint depth-scale adjustment (JDSA)-coupled dense bundle adjustment estimates poses and scale-consistent inverse depth, which are back-projected into 3D Gaussian ellipsoids; keyframes are grouped into bounded submaps, and intra-agent loops trigger online \(\mathrm{Sim}(3)\) pose-graph bundle adjustment (PGBA).
    \textbf{(2) Global intra-to-inter loop closure}: every frozen submap is encoded as a compact summary, a 128-D descriptor, sparse 3D feature points, a downsampled registration cloud, a bounding volume, and an anchor keyframe. Cosine top-\(k\) retrieval, hybrid sparse and LoFTR matching with RANSAC/Umeyama \(\mathrm{Sim}(3)\), a spatial-extent gate, and ICP refinement produce verified inter-agent constraints, which are optimized jointly with intra-agent temporal edges and Sim(3) multi-agent pose-graph bundle adjustment (MaPGBA) in a global \(\mathrm{Sim}(3)\) submap pose graph coupling geometric \(r^{\text{geo}}\) and anchor-image photometric \(r^{\text{pho}}\) residuals.
    \textbf{(3) Gaussian fusion and global refinement}: an occupancy-grid test deduplicates borrowed Gaussians against each target agent's occupied and free-space voxels, and a final joint photometric refinement of the fused Gaussians, keyframe poses, and exposures produces the globally consistent photorealistic map. Each submap advances through the state flow ACTIVE\,\(\rightarrow\)\,FROZEN\,\(\rightarrow\)\,VERIFIED\,\(\rightarrow\)\,ALIGNED\,\(\rightarrow\)\,FUSABLE\,\(\rightarrow\)\,FUSED.
    }
  \label{fig:pipeline}
  \vspace{-0.8\baselineskip}
\end{figure*}

\section{Related Work}
\label{sec:related}

\subsection{Dense and Photorealistic Visual SLAM}
Visual SLAM systems have progressively expanded from accurate camera tracking toward dense, photorealistic scene reconstruction. 
Sparse tracking systems such as ORB-SLAM3~\cite{campos2021orb} and DSO~\cite{engel2017direct} 
provide accurate camera poses but limited scene geometry, 
while dense or semi-dense direct methods such as DTAM~\cite{newcombe2011dtam} 
and LSD-SLAM~\cite{engel2014lsd} improve reconstruction density. 
Recent learning-based systems, including DROID-SLAM~\cite{teed2021droid}, 
DPVO~\cite{teed2023deep}, and MASt3R-SLAM~\cite{murai2025mast3r}, 
further improve correspondence estimation, bundle adjustment, 
and monocular reconstruction with learned priors. {A parallel line of work replaces incremental estimation by feed-forward
3-D foundation models: DUSt3R~\cite{wang2024dust3r} and
MASt3R~\cite{leroy2024grounding} regress pointmaps directly from image pairs,
MASt3R-SfM~\cite{duisterhof2025mast3r}, Spann3R~\cite{wang20253d},
CUT3R~\cite{wang2025continuous}, and SLAM3R~\cite{liu2025slam3r} extend them to
video, and VGGT~\cite{wang2025vggt} and VGGT-SLAM~\cite{maggio2026vggt} scale
the idea to whole sequences. Such priors markedly improve monocular robustness,
but most remain single-agent and leave the cross-agent scale and alignment
problem studied here untouched.}
Neural implicit SLAM methods~\cite{sucar2021imap, zhu2022nice, zhu2024nicer, yang2022vox, wang2023co, johari2023eslam, sandstrom2023point, rosinol2023nerf, zhang2023go, liso2024loopy, zhang2023hi} 
enable high-fidelity geometry and appearance reconstruction, 
but volume rendering and neural optimization limit real-time scalability. 
3D Gaussian Splatting (3DGS)~\cite{kerbl20233d} and surface-oriented variants such as 2DGS~\cite{huang20242d} 
provide explicit differentiable primitives with efficient rasterization 
and have been adopted in single-agent Gaussian SLAM~\cite{keetha2024splatam, yan2024gs, matsuki2024gaussian, zhang2025hi, yugay2023gaussian, ha2024rgbd, peng2024rtg, li2024sgs, hu2024cg, zhu2025loopsplat, sandstrom2025splat}. 
These methods achieve photorealistic reconstruction from monocular or RGB-D input, 
but do not address inter-agent coordination, cross-agent scale alignment, 
or duplicate-aware Gaussian fusion.

\subsection{Multi-Agent Collaborative SLAM}
Multi-agent collaborative SLAM addresses single-agent coverage limitations 
through parallel exploration, inter-agent loop closure, and submap sharing. 
Sparse collaborative frameworks~\cite{schmuck2019ccm, schmuck2021covins, lajoie2020door, tian2022kimera, lajoie2023swarm} 
support trajectory alignment and distributed optimization, 
but are not designed for photorealistic rendering. 
Neural-field extensions~\cite{hu2023cp, deng2025mne, deng2025mcn} reduce communication overhead, 
yet inherit the rendering and scalability limits of implicit representations.

Recent 3DGS-based collaborative systems~\cite{yugay2025magic, xu2025mac, thomas2025grand, chen2026coma, yu2025hammer} 
enable renderable maps, but typically rely on RGB-D input, 
where metric depth simplifies inter-agent alignment to rigid \(\mathrm{SE}(3)\). 
In monocular settings, independently optimized agents can produce inconsistent scales, 
requiring \(\mathrm{Sim}(3)\) alignment and duplicate-aware fusion 
to avoid redundant Gaussians and density inconsistencies.

Among RGB-only baselines, the MultiSLAM-DiffPose~\cite{lipson2024multi} 
yields only sparse geometry, the MANG-SLAM~\cite{li2025mang} treats 3DGS as a downstream renderer guided by NeRF submap priors rather than a native representation, 
and the MA-MASt3R-SLAM~\cite{zhou2026multi} performs dense monocular map fusion 
with 3D reconstruction priors but does not exploit the explicit and differentiable renderable representation of 3DGS. 
Monocular RGB-only collaborative 3DGS SLAM with scale-consistent \(\mathrm{Sim}(3)\) alignment 
and Gaussian fusion therefore remains underexplored.

\section{MAGS-SLAM}
\label{sec:method}
Our MAGS-SLAM framework is a monocular RGB-only multi-agent 3D Gaussian Splatting SLAM system (Fig.~\ref{fig:pipeline}).
Each agent runs an independent monocular front-end that builds local Gaussian submaps,
while a lightweight coordinator discovers overlaps from compact submap summaries,
maintains a global \(\mathrm{Sim}(3)\) pose graph, and performs occupancy-aware fusion. {The system requires no inter-agent extrinsic calibration and no common reference frame; every cross-agent relation is recovered online from imagery. The pinhole intrinsics of each camera are assumed known.}
Accordingly, the rest of this section is organized into three components:
per-agent local monocular Gaussian SLAM (Section \ref{sec:intra_agent}),
the global collaborative \(\mathrm{Sim}(3)\) pose graph maintained by the coordinator (Section \ref{sec:inter_agent}),
and occupancy-aware asynchronous fusion of the aligned submaps (Section \ref{sec:gs_fusion}).

\subsection{Local Per-Agent Monocular Gaussian SLAM}
\label{sec:intra_agent}

Each agent runs an independent monocular DROID-based~\cite{teed2021droid} front-end that couples recurrent dense bundle adjustment (BA) with a learned monocular depth prior~\cite{hu2024metric3d}.
Over a sliding co-visibility window, the poses \(\{\mathbf{T}\}\) and depth maps \(\{\mathbf{d}\}\) of the keyframe graph \(\mathcal{G} = (\mathcal{V}, \mathcal{E})\) are refined by minimizing the reprojection cost
\begin{equation}
\mathcal{L}^{\text{BA}} = \sum_{(i,j)\in\mathcal{E}}
\bigl\|\mathbf{u}^*_{ij} -
\pi\bigl(\mathbf{T}_{ij} \pi^{-1}(\mathbf{u}_i, \mathbf{d}_i)\bigr)
\bigr\|^2_{\bm{\Sigma}_{ij}},
\label{eq:dense_ba}
\end{equation}
where \(\mathbf{u}^*_{ij}\) is the refined correspondence~\cite{teed2021droid} and
\(\pi,\pi^{-1}\) are the camera projection and back-projection operators. Since a monocular front-end cannot fix metric scale, we additionally couple per-keyframe scale coefficients \(\{\mathbf{s}\}\) and depth maps \(\{\mathbf{d}\}\) through Joint Depth-Scale Adjustment (JDSA)~\cite{zhang2023hi, zhang2025hi, cao2025mcgs},
\begin{equation}
\mathcal{L}^{\text{JDSA}} = \mathcal{L}^{\text{BA}}
+ \sum_{k\in \mathcal{V}}
\bigl\| \mathcal{B}_k\mathbf{d}_k - \mathbf{d}^{\text{prior}}_k \bigr\|^2,
\label{eq:jdsa}
\end{equation}
where \(\mathbf{d}^{\text{prior}}_k\) is the learned prior from Metric3D~\cite{hu2024metric3d} and \(\mathcal{B}_k(\mathbf{u}_k, \mathbf{s}_k)\) bilinearly interpolates the four surrounding scale coefficients to rescale the depth. {Keyframe selection and loop-candidate detection use optical-flow, covisibility, temporal-gap and rotation thresholds listed in Table~\ref{tab:params}.} When the loop detector reports intra-agent loop-closure candidates, we run online \(\mathrm{Sim}(3)\) pose-graph bundle adjustment (PGBA)~\cite{zhu2026vigs, zhang2025hi} over the full keyframe history to correct both short- and long-term drift,
\begin{equation}
\mathcal{L}^{\text{PGBA}} = \mathcal{L}^{\text{BA}^*}
+ \sum_{(i,j)\in\mathcal{E}^+}
\bigl\| \text{log}\bigl( \hat{\mathbf{T}}_{ij}\mathbf{T}_{i}\mathbf{T}_{j}^{-1} \bigr) \bigr\|^2_{\bm{\Sigma}^+_{ij}}.
\label{eq:pgba}
\end{equation}
Here, \(\mathcal{L}^{\text{BA}^*}\) is with \(\mathcal{G}^*\), where \(\mathcal{E}^*\) in \(\mathcal{G}^*\) is the loop edges based on optical flow differences. \(\mathcal{E}^+\) is the relative pose graph by adding the relative pose from graph \(\mathcal{G}\). \(\hat{\mathbf{T}}_{ij}\) is the relative pose in the pose graph. Equation~\eqref{eq:pgba} is an intra-agent operator: it is solved by each agent over its own keyframes and entirely in its own gauge; Section \ref{sec:mapgba} lifts the same construction one level up to a submap graph shared by the whole team, which we refer to as multi-agent PGBA (MaPGBA). When a new keyframe is registered, its inverse-depth map is back-projected to seed new Gaussians.
Each Gaussian is initialized with center \(\bm{\mu} = \mathbf{T}_k^{-1}\pi^{-1}(\mathbf{u}, \mathbf{d}_k)\), 
color from the pixel, and scale from the local neighborhood.
The local Gaussian map is then optimized against a sliding window of keyframes by gradient descent on
\begin{equation}
\mathcal{L}^{\text{map}} =
\alpha\mathcal{L}_1(\hat{I}, I)
+ \beta\mathcal{L}_1(\hat{D}, D)
+ \lambda_n \mathcal{L}_{\text{normal}}
+ \lambda_s \mathcal{L}_{\text{iso}},
\label{eq:map_loss}
\end{equation}
with {local-mapping} weights \(\alpha=0.95\), \(\beta=0.25\), \(\lambda_n=10^{-2}\), \(\lambda_s=10\).
The photometric and depth terms reduce to per-keyframe L1 errors over the sliding window \(\mathcal{K}\), where \(D\) denotes the BA-estimated depth used as self-supervision.
The normal term \(\mathcal{L}_{\text{normal}}=\sum_{k\in\mathcal{K}}\bigl|1-\hat{N}_k^{\top}\bar{N}_k\bigr|\) is a cosine embedding loss between the depth-induced normal \(\hat{N}_k\) and the monocular normal prior \(\bar{N}_k\), and the regularizer \(\mathcal{L}_{\text{iso}}=\sum_{i\in\mathcal{G}}|\mathbf{s}_i-\bar{s}_i|\) pulls each Gaussian's scale components toward their mean \(\bar{s}_i\) to penalize ellipsoid stretching. The optimization above runs continuously over a sliding keyframe window. To interface with the inter-agent coordinator, we additionally bundle the keyframe stream into bounded submaps \(\mathcal{P}^a_\ell\) (indexed by agent \(a\in\mathcal{A}\) and per-agent submap counter \(\ell=1,\dots,L_a\); a second submap is denoted \((b,m)\)), each holding its keyframes' poses, depths, and Gaussians. A submap is frozen once it reaches \(K_{\max}\) keyframes or the camera centers exceed a Euclidean distance of \(\tau_{\text{move}}\); freezing detaches it from local bundle adjustment and assigns it a per-submap \(\mathrm{Sim}(3)\) correction \(C^a_\ell\in\mathrm{Sim}(3)\) that maps the submap's local coordinates into the aligned global frame, initialized to \(C^a_\ell=\mathbf{I}\) and later refined jointly by the coordinator. Successive frozen submaps from the same agent are linked by a temporal \(\mathrm{Sim}(3)\) edge \(e\in\mathcal{E}^t\) of the global submap graph \(\mathcal{G}^{\text{sm}}\) (Section~\ref{sec:inter_agent}). The above chains anchor the looser inter-agent constraints.

\subsection{Global Collaborative Pose Graph Optimization}
\label{sec:inter_agent}
The coordinator turns each agent's frozen submaps into a global \(\mathrm{Sim}(3)\) graph in four stages, including \textbf{encoding}, \textbf{retrieval}, \textbf{verification}, \textbf{joint optimization}, augmented by a reactive path that integrates PGBA corrections without re-running the front-end. Together with the per-agent PGBA of Eq.~\eqref{eq:pgba}, this graph forms the multi-agent PGBA (MaPGBA) problem that we formalize in Section \ref{sec:mapgba}.

\subsubsection{Compact submap summary}
Raw keyframes and Gaussians are too large for inter-agent transmission, so each frozen submap is encoded as
\begin{equation}
\mathcal{S}^a_\ell = \bigl(\mathbf{d}^a_\ell,\;
\mathcal{Q}^a_\ell,\; \mathcal{R}^a_\ell,\;
[\mathbf{b}_{\min}, \mathbf{b}_{\max}]^a_\ell,\;
\mathbf{A}^a_\ell \bigr),
\end{equation}
where \(\mathbf{d}^a_\ell\in\mathbb{R}^{128}\) is a mean-pooled DROID global descriptor, \(\mathcal{Q}^a_\ell\) holds up to \(N_{\mathcal{Q}}\) salient 3D points with local descriptors selected by the per-pixel score
\begin{equation}
\sigma(\mathbf{u}) = \|\nabla\mathbf{F}\|_{\mathbf{u}}
+ \lambda_{d}\|\nabla\mathbf{d}\|_{\mathbf{u}}
+ \lambda_{F}\|\mathbf{F}\|_{\mathbf{u}},
\end{equation}
with \(\mathbf{F}\) the per-pixel DROID feature{, \(\nabla\mathbf{d}\) taken on the keyframe inverse-depth map, and \(\lambda_d=0.35\), \(\lambda_F=0.10\)}; the three terms reward feature-edge strength, depth boundaries, and feature distinctiveness. \(\mathcal{R}^a_\ell\) is a registration cloud for ICP \cite{besl1992method}, the bounding volume \([\mathbf{b}_{\min},\mathbf{b}_{\max}]^a_\ell\) supplies the normalization scale of Eq.~\eqref{eq:extent_gate}, and \(\mathbf{A}^a_\ell\) packages the anchor keyframe (pose, image, disparity, intrinsics).

\subsubsection{Retrieval and correspondence}
Each new summary is matched against a top-\(K\) shortlist \(\{\mathcal{S}^b_m\}\) by cosine similarity on \(\mathbf{d}^a_\ell\) gated by \(\tau_{\text{sim}}\). For each candidate \(\mathcal{S}^b_m\), a matcher \(\Pi(\mathcal{S}^a_\ell,\mathcal{S}^b_m;\bm{\theta}_{\text{match}})\), including mutual nearest-neighbor matching jointly gated by an absolute cosine threshold, a best/second-best margin, and Lowe's ratio test bundled into \(\bm{\theta}_{\text{match}}\), returns a correspondence set \(\mathcal{M}\) with \(|\mathcal{M}|\ge N_{\min}\). When texture is sparse, dense LoFTR~\cite{sun2021loftr} matches between anchor images are back-projected and appended to the set.

\subsubsection{Sim(3) estimation and verification cascade}
\label{sec:sim3_estimation}
A RANSAC loop estimates the relative \(\mathrm{Sim}(3)\) transform \(T^{t\leftarrow s}\) via Umeyama~\cite{umeyama1991least}:
\begin{equation}
\bm{\Gamma} = \tfrac{1}{N}\sum_i (\mathbf{p}^t_i-\bar{\mathbf{p}}^t)(\mathbf{p}^s_i-\bar{\mathbf{p}}^s)^\top,\quad
\bm{\Gamma} = \mathbf{U}\bm{\Lambda}\mathbf{V}^\top,
\label{eq:umeyama_svd}
\end{equation}
\begin{equation}
\mathbf{R} = \mathbf{U}\mathbf{D}\mathbf{V}^\top,\quad
s = \tfrac{\operatorname{tr}(\bm{\Lambda}\mathbf{D})}{\sigma_s^2},\quad
\mathbf{t} = \bar{\mathbf{p}}^t - s\mathbf{R}\bar{\mathbf{p}}^s,
\label{eq:umeyama_recover}
\end{equation}
with \(\bm{\Lambda}\) the diagonal of singular values of \(\bm{\Gamma}\), \(\mathbf{D}\) handling reflections, and \(\sigma_s^2=\tfrac{1}{N}\sum_i\|\mathbf{p}^s_i-\bar{\mathbf{p}}^s\|^2\) the source-point variance. The surviving inlier set is denoted \(\mathcal{X}_{\text{inl}}=\mathcal{X}^s_{\text{inl}}\cup\mathcal{X}^t_{\text{inl}}\), with \(\mathcal{X}^{s}_{\text{inl}}\) and \(\mathcal{X}^{t}_{\text{inl}}\) the matched 3D points in the source and target submap (\(\mathcal{P}^s\), \(\mathcal{P}^t\)) respectively, and is then refined by ICP~\cite{besl1992method}. Rather than gating each candidate against a stack of independent thresholds, we collect all per-pair checks into a single verification predicate with the bundled gate thresholds \(\bm{\theta}_V\),
\begin{equation}
V\bigl(s,\Phi,\rho_{\text{icp}},\eta;\bm{\theta}_V\bigr) \in \{0,1\},
\label{eq:cascade}
\end{equation}
which is true iff the recovered scale \(s\) lies in the monocular plausibility band, the ICP fitness \(\Phi\) and RMSE \(\rho_{\text{icp}}\) clear their bounds, and the spatial-extent ratio
\begin{equation}
\eta = \min\Bigl(\tfrac{\mathrm{ext}(\mathcal{X}^s_{\text{inl}})}{\mathrm{diag}(\mathcal{P}^s)},
\tfrac{\mathrm{ext}(\mathcal{X}^t_{\text{inl}})}{\mathrm{diag}(\mathcal{P}^t)}\Bigr) \geq \tau_{\text{ext}},
\label{eq:extent_gate}
\end{equation}
clears the gate; here \(\mathrm{ext}(\mathcal{X})=\|\max\mathcal{X}-\min\mathcal{X}\|\) is the AABB diagonal of a point set and \(\mathrm{diag}(\mathcal{P}^{s/t})=\|\mathbf{b}^{s/t}_{\max}-\mathbf{b}^{s/t}_{\min}\|\) reuses the submap bounding box from \(\mathcal{S}\).

\begin{algorithm}[t]
\SetAlgoLined
\DontPrintSemicolon
\caption{Submap Graph Update}
\label{alg:coordinator}
\KwIn{summary \(\mathcal{S}^a_\ell\); PGBA fit \((\Delta_{a,\ell},\rho_{\text{rig}})\)}
\KwOut{updated corrections \(\{C^a_\ell\}\), fusable set \(\mathcal{F}\)}
\BlankLine
\eIf{\(\mathcal{S}^a_\ell\) is new}{
   add edge \(\mathcal{S}^a_{\ell-1}\to\mathcal{S}^a_\ell\) to \(\mathcal{E}^t\) with weight \(w^t\)\;
   \ForEach{\(\mathcal{S}^b_m \in \mathrm{TopK}(\mathbf{d}^a_\ell;\tau_{\text{sim}})\)}{
      \(\mathcal{M}\leftarrow\Pi(\mathcal{S}^a_\ell,\mathcal{S}^b_m;\bm{\theta}_{\text{match}})\)\;
      \(T^{t\leftarrow s},\mathcal{X}_{\text{inl}}\leftarrow\text{Ransac-Umeyama}(\mathcal{M})\)\;
      \(T^{t\leftarrow s}\leftarrow\textsc{Icp}(T^{t\leftarrow s},\mathcal{R}^a_\ell,\mathcal{R}^b_m)\)\;
      \lIf{\(V(\cdot;\bm{\theta}_V)\) \rm(Eq.~\eqref{eq:cascade})}
          {add edge \((T^{t\leftarrow s},w^v)\) to \(\mathcal{E}^v\)}
   }
}{
   \eIf{\(\rho_{\text{rig}}>\tau_{\text{rig}}\)}
       {invalidate inter-agent edges incident to \(\mathcal{S}^a_\ell\)}
       {update each \(M_e\) analytically (Eq.~\eqref{eq:edge_update})}
   drop photometric cache of \(\mathcal{S}^a_\ell\)\;
}
\BlankLine
\(\{C^a_\ell\}\leftarrow\arg\min \mathcal{L}^{\text{graph}}\) (Eq.~\eqref{eq:graph_cost})\;
\(\mathcal{F}\leftarrow\{(a,\ell):\bar{\rho}_e(\mathcal{S}^a_\ell)\le\tau_{\text{res}} \wedge |\mathcal{E}^v(\mathcal{S}^a_\ell)|\ge1\}\)\;
\end{algorithm}

\begin{figure*}[!t]
    \centering
    \includegraphics[width=\textwidth]{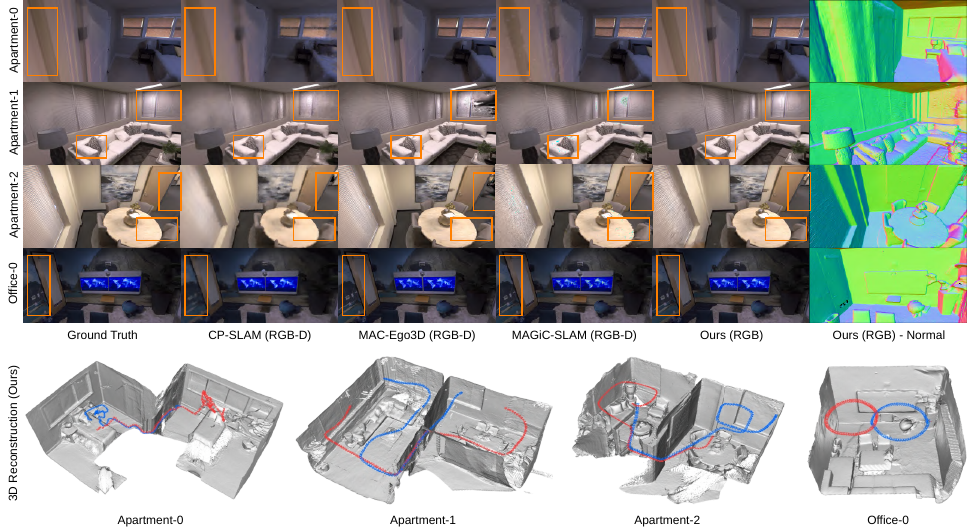}
    \vspace{-1.4\baselineskip}
    \caption{\textbf{Qualitative comparison on the ReplicaMultiagent dataset.}
    \emph{Top:} rendered views on four scenes (Apartment-0/1/2 and Office-0). From left to right we show the ground-truth view, three RGB-D baselines (CP-SLAM~\cite{hu2023cp}, MAC-Ego3D~\cite{xu2025mac}, MAGiC-SLAM~\cite{yugay2025magic}), our RGB-only result, and our rendered surface normals; the coloured boxes mark zoom-in regions. Although MAGS-SLAM uses no depth input, it recovers sharper textures and thinner structures with fewer artefacts, and its normal maps confirm geometrically consistent surfaces.
    \emph{Bottom:} the fused multi-agent 3D reconstruction of MAGS-SLAM for each scene, with the two per-agent camera trajectories overlaid (red and blue). The maps remain seamless across agents, without visible duplication or misalignment at the overlap boundaries.}
    \label{fig:replica_results}
    \vspace{-1.4\baselineskip}
\end{figure*}

\subsubsection{Pose graph optimization objective}
The accepted constraints from Section \ref{sec:sim3_estimation} feed a \(\mathrm{Sim}(3)\) submap pose graph \(\mathcal{G}^{\text{sm}}=(\{\mathcal{P}^a_\ell\},\mathcal{E}^{\text{sm}})\) with \(\mathcal{E}^{\text{sm}}=\mathcal{E}^t\cup\mathcal{E}^v\) (intra-agent temporal and verified inter-agent edges), where the \(\mathrm{Sim}(3)\) parameterization is used to accommodate the per-agent monocular scale ambiguity. We optimize the \(\mathrm{Sim}(3)\) parameterization with the cost couples geometric residuals
\begin{equation}
\mathbf{r}^{\text{geo}}_e = \log\bigl(M_e^{-1} \circ
C_{\text{tgt}}^{-1} \circ C_{\text{src}}\bigr) \in \mathbb{R}^7
\label{eq:geo_residual}
\end{equation}
and anchor-image photometric residuals
\begin{equation}
\mathbf{r}^{\text{pho}}_e(\mathbf{u}) = I^{\text{src}}(\mathbf{u})
- I^{\text{tgt}}\bigl(
\pi^{\text{tgt}}(C_{\text{tgt}}^{-1}\circ C_{\text{src}}\cdot
\mathbf{p}^{\text{src}}(\mathbf{u}))\bigr),
\label{eq:pho_residual}
\end{equation}
where \(M_e\!\in\!\mathrm{Sim}(3)\) is the verified source to target relative transform on edge \(e\), \(C_{\text{src}},C_{\text{tgt}}\) are the per-submap corrections of its endpoints, and \(I^{\text{tgt}}(\cdot)\) denotes bilinear sampling of the target anchor image at sub-pixel coordinates. Jacobians w.r.t.\ \(C_{\text{src}},C_{\text{tgt}}\) follow by chain rule through this image sampler and \(\pi^{\text{tgt}}\), as in standard direct alignment on \(\mathrm{Sim}(3)\)~\cite{engel2017direct, engel2014lsd}. Here, the jointly geometric and photometric optimization objective is
\begin{equation}
\mathcal{L}^{\text{graph}}\bigl(\{C^a_\ell\}\bigr) =
\sum_{e\in\mathcal{E}^{\text{sm}}} w_e \bigl\|\mathbf{r}^{\text{geo}}_e\bigr\|^2_{\rho}
+ \sum_{e\in\mathcal{E}^{v}} \frac{w^{\text{pho}}_e}{|\mathcal{U}_e|}
  \bigl\|\mathbf{r}^{\text{pho}}_e\bigr\|^2_{\rho},
\label{eq:graph_cost}
\end{equation}
where \(\{C^a_\ell\}=\{C^a_\ell : a\in\mathcal{A},\ell=1,\dots,L_a\}\) is the set of all per-submap corrections and \(\mathcal{U}_e\) is the in-frame, finite-depth anchor-pixel set of edge \(e\); the per-edge normalization makes inter-agent pairs of different overlap extents comparable{, and \(\|\cdot\|^2_{\rho}\) is the Huber loss with unit threshold}. Edge weights split into intra-agent temporal \(w^t\) (on \(\mathcal{E}^t\)) and inter-agent \(w^v\) (on \(\mathcal{E}^v\)) with {\(w^t>w^v\) (\(w^t=5.0\) and \(w^v=1.0\), Table~\ref{tab:params})}, so trusted temporal chains anchor the looser inter-agent constraints; \(w^{\text{pho}}_e\) only fires on verified inter-agent edges. The graph is solved by damped Gauss-Newton within a Levenberg-Marquardt trust region~\cite{nocedal2006numerical, levenberg1944method}, with the first frozen submap fixed at identity. After convergence, only submaps whose mean per-edge residual \(\bar{\rho}_e(\mathcal{S}^a_\ell)\) falls below \(\tau_{\text{res}}\) and that carry at least one inter-agent edge are forwarded to fusion (Section \ref{sec:gs_fusion}); the rest stay in the graph for re-evaluation as new evidence arrives.

\subsubsection{Multi-agent PGBA (MaPGBA)}
\label{sec:mapgba}
Equations~\eqref{eq:pgba} and \eqref{eq:graph_cost} instantiate the same problem class at two levels of granularity. Intra-agent PGBA refines a keyframe graph of a single agent by combining a bundle-adjustment data term \(\mathcal{L}^{\text{BA}^*}\) with relative-pose terms on the loop-augmented edge set \(\mathcal{E}^+\); the coordinator refines a submap graph shared by all agents by {combining relative-\(\mathrm{Sim}(3)\) terms on \(\mathcal{E}^{\text{sm}}=\mathcal{E}^t\cup\mathcal{E}^v\) with the anchor-image photometric term \(\mathbf{r}^{\text{pho}}_e\), which serves as the bundle-adjustment data term when the raw correlation features of a remote agent are unavailable.}
{Specifically, anchor pixels are back-projected using the anchor inverse depth, transformed by the current submap corrections, and re-projected into the target view, where target intensities are re-sampled at every iteration. The resulting Jacobians therefore propagate through the target projection and the transformed 3-D anchor points, yielding the same photometric reprojection mechanism as structure-fixed direct bundle adjustment. Thus, Eq.~\eqref{eq:graph_cost} can be viewed as a structure-fixed Sim(3) bundle adjustment that is fully evaluable from the transmitted summaries alone.}
We therefore treat the two levels as one estimation problem, multi-agent pose-graph bundle adjustment (MaPGBA),
\begin{equation}
\mathcal{L}^{\text{MaPGBA}} =
\sum_{a\in\mathcal{A}} \mathcal{L}^{\text{PGBA}}_a\bigl(\{\mathbf{T}\}^a\!,\{\mathbf{d}\}^a\bigr)
+ \mathcal{L}^{\text{graph}}\bigl(\{C^a_\ell\} \bigr),
\label{eq:mapgba}
\end{equation}
where \(\{\mathbf{T}\}^a,\{\mathbf{d}\}^a\) are the keyframe poses and depths of agent \(a\) expressed in its own local frame. {The coordinator term in Eq.~\eqref{eq:graph_cost} is evaluated using quantities derived from these agent states, namely the verified measurements \(M_e\) and the anchor payloads entering \(\mathbf{r}^{\text{pho}}_e\), which mediate the coupling between the two levels.} The two levels are tied by the lifting relation
\begin{equation}
\mathbf{T}^{\text{glob}}_k = \mathbf{T}_k \circ \bigl(C^a_\ell\bigr)^{-1},
\quad
\bm{\mu}^{\text{glob}}_i = C^a_\ell \cdot \bm{\mu}_i,
\quad k,i \in \mathcal{P}^a_\ell,
\label{eq:mapgba_lift}
\end{equation}
which carries every keyframe pose and Gaussian center of submap \(\mathcal{P}^a_\ell\) into the aligned global frame; \(\mathbf{T}_k\) is a world-to-camera pose, so the correction acts on its right, and Eq.~\eqref{eq:gauss_update} gives the induced action on a full Gaussian. Both levels are parameterized on \(\mathrm{Sim}(3)\): the scale block of \(\mathcal{L}^{\text{PGBA}}_a\) absorbs the drift of agent \(a\)'s own monocular scale, while the scale block of \(\{C^a_\ell\}\) reconciles the mutually incompatible scales of different agents.

Minimizing Eq.~\eqref{eq:mapgba} jointly over \(\{\mathbf{T}\},\{\mathbf{d}\}\) and \(\{C^a_\ell\}\) would require the coordinator to hold every agent's keyframes and dense front-end features: agents transmit only the summaries \(\mathcal{S}^a_\ell\) and run asynchronously. We therefore minimize Eq.~\eqref{eq:mapgba} with an alternating block-coordinate scheme,
\begin{equation}
\begin{aligned}
&\{\{\mathbf{T}\},\{\mathbf{d}\}\}^{a,(n+1)} = \arg\min \; \mathcal{L}^{\text{PGBA}}_a, \quad \forall a\in\mathcal{A},\\
&\{C^a_\ell\}^{(n+1)} = \arg\min \; \mathcal{L}^{\text{graph}}\bigl(\cdot \,\big|\, \{\{\mathbf{T}\},\{\mathbf{d}\}\}^{(n+1)}\bigr).
\end{aligned}
\label{eq:mapgba_bcd}
\end{equation}
{Eq.~\eqref{eq:mapgba_bcd} implements a block-alternating optimization
scheme for Eq.~\eqref{eq:mapgba} with structured coupling between the agent
and coordinator blocks. Each agent updates its local variables
\((\{\mathbf{T}\}^a,\{\mathbf{d}\}^a)\) independently, while the coordinator
optimizes \(\{C^a_\ell\}\); their interaction is carried through the
inter-agent measurements \(M_e\) and anchor payloads. This structure permits
all agent blocks to be optimized in parallel, followed by a coordinator
update. Between coordinator solves, each sufficiently rigid submap is
represented by a single similarity increment \(\Delta_{a,\ell}\), and
Eq.~\eqref{eq:edge_update}, together with the corresponding gauge update,
transports its associated measurements into the updated local frame. In the
ideal case \(\rho_{\mathrm{rig}}=0\), this representation is exact: the agent
update, gauge change, and measurement transport constitute a consistent
change of variables between the agent and coordinator representations.
Accordingly, the underlying inter-agent measurement geometry is preserved,
target-side geometric residuals remain invariant, and zero source-side
residuals remain zero. The agent block can therefore reduce its local
\(\mathcal{L}^{\mathrm{PGBA}}_a\) without introducing an inconsistency into
the coordinator term, giving the block-coordinate interpretation of
Eq.~\eqref{eq:mapgba_bcd}. For
\(0<\rho_{\mathrm{rig}}\le\tau_{\mathrm{rig}}\), this equivalence is only
approximate, and the rigidity test bounds the departure from the
single-similarity model. Updates violating this condition are instead
re-verified before subsequent optimization. Thus, in practice,
Eq.~\eqref{eq:mapgba_bcd} is an approximate block-coordinate scheme whose
validity is controlled by \(\rho_{\mathrm{rig}}\).}
Therefore, keeping the alternation in Eq.~\eqref{eq:mapgba_bcd} well posed requires transporting the coordinator's data into the new local frame after every agent-block step. We call this requirement MaPGBA consistency and realize it as the reactive path of Section \ref{sec:reactive}.

\begin{figure*}[!t]
    \centering
    \includegraphics[width=\textwidth]{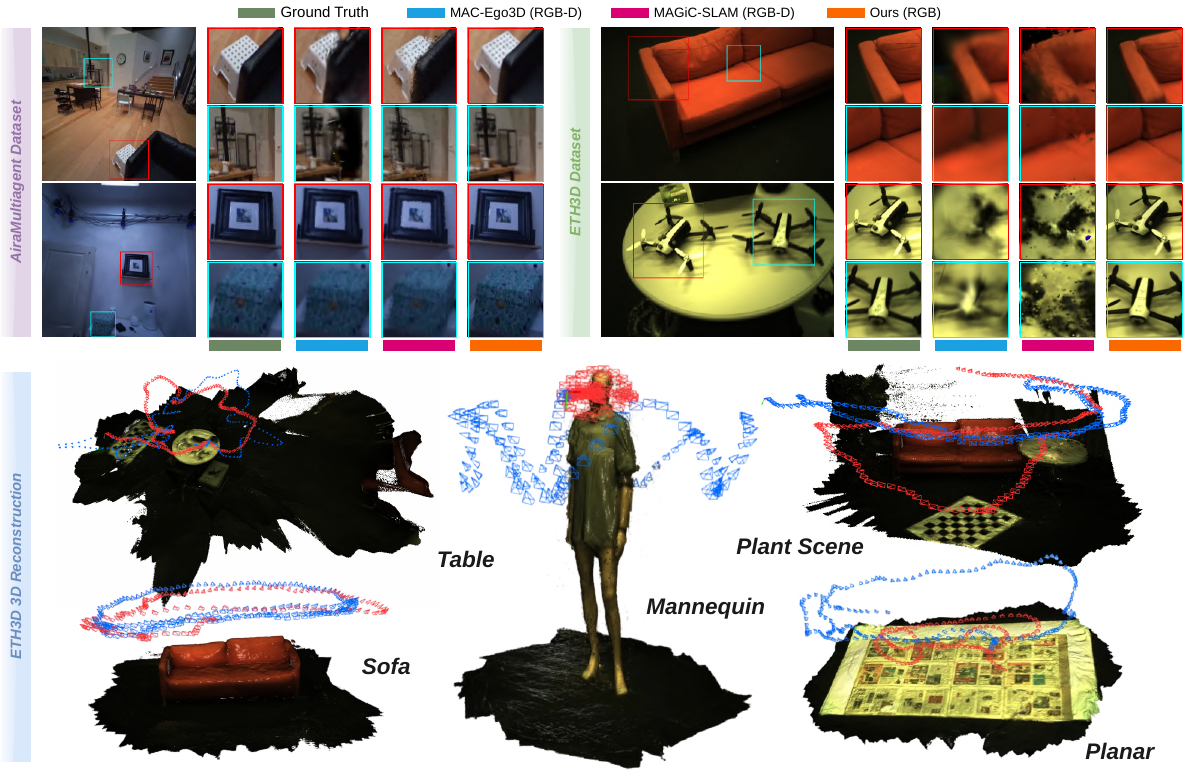}
    \vspace{-1.7\baselineskip}
    \caption{\textbf{Qualitative comparison on the real-world AriaMultiagent~\cite{pan2023aria} and ETH3D~\cite{schops2019bad} datasets.}
    \emph{Top:} renderings with zoom-in insets on AriaMultiagent (\emph{left}) and ETH3D (\emph{right}); within each inset we compare the ground truth against the RGB-D baselines MAC-Ego3D~\cite{xu2025mac} and MAGiC-SLAM~\cite{yugay2025magic} and our RGB-only reconstruction (border colours match the legend). Under real sensor noise and challenging illumination, MAGS-SLAM preserves fine detail and colour fidelity where the RGB-D baselines blur or drop thin structures.
    \emph{Bottom:} the fused ETH3D reconstructions produced by MAGS-SLAM for the Sofa, Table, Plant-Scene, Mannequin, and Planar sequences, with the per-agent trajectories (red and blue) overlaid to show consistent cross-agent alignment.}
    \label{fig:eth3d_aria_results}
    \vspace{-0.7\baselineskip}
\end{figure*}

\begin{table*}[t]
\caption{
    Multi-agent tracking accuracy comparison (ATE RMSE [cm] $\downarrow$) on ReplicaMultiagent datasets. Best results are highlighted as \colorbox{colorFst}{\textbf{first}}, \colorbox{colorSnd}{second}, and \colorbox{colorTrd}{third}. "--" indicates sequences without reported results in the original papers or unavailable code. "\ding{55}" indicates invalid results due to the failure.
    }
\vspace{-0.3cm}
\label{tab:tracking_replica}
\setlength{\tabcolsep}{4pt}
\centering
\footnotesize
\begin{tabular}{l c | ccc | ccc | ccc | ccc}
\toprule
\multirow{2}{*}{\textbf{Method}} &
\multirow{2}{*}{\textbf{Input}} &
\multicolumn{3}{c|}{\textbf{Office-0}} &
\multicolumn{3}{c|}{\textbf{Apartment-0}} &
\multicolumn{3}{c|}{\textbf{Apartment-1}} &
\multicolumn{3}{c}{\textbf{Apartment-2}} \\
& & Agent 1 & Agent 2 & Avg. & Agent 1 & Agent 2 & Avg. & Agent 1 & Agent 2 & Avg. & Agent 1 & Agent 2 & Avg. \\
\midrule
Swarm-SLAM~\cite{lajoie2023swarm}               & RGB-D & 1.07 & 1.76 & 1.42 & 1.61 & 1.98 & 1.80 & 4.62 & 6.50 & 5.56 & 2.69 & 8.53 & 5.61 \\
CP-SLAM~\cite{hu2023cp}                    & RGB-D & 0.50 & 0.79 & 0.65 & 0.62 & 1.28 & 0.95 & 1.11 & 1.72 & 1.42 & 1.41 & 2.41 & 1.91 \\
MNE-SLAM~\cite{deng2025mne}                  & RGB-D & \cellcolor{colorFst}\textbf{0.28} & \cellcolor{colorTrd}0.33 & \cellcolor{colorTrd}0.31 & \cellcolor{colorTrd}0.43 & 0.54 & 0.48 & 1.21 & 0.99 & 1.10 & 0.43 & 0.74 & 0.59 \\
MCN-SLAM~\cite{deng2025mcn}                  & RGB-D & -- & -- & -- & \cellcolor{colorSnd}0.42 & 0.52 & \cellcolor{colorSnd}0.46 & 1.18 & 0.97 & 1.07 & \cellcolor{colorTrd}0.41 & 0.71 & 0.57 \\
MAGiC-SLAM~\cite{yugay2025magic}              & RGB-D & \cellcolor{colorSnd}0.31 & \cellcolor{colorSnd}0.24 & \cellcolor{colorFst}\textbf{0.27} & \cellcolor{colorFst}\textbf{0.13} & \cellcolor{colorFst}\textbf{0.21} & \cellcolor{colorFst}\textbf{0.16} & \cellcolor{colorFst}\textbf{0.21} & \cellcolor{colorFst}\textbf{0.30} & \cellcolor{colorFst}\textbf{0.26} & 0.42 & \cellcolor{colorFst}\textbf{0.22} & \cellcolor{colorSnd}0.32 \\
\midrule
CCM-SLAM~\cite{schmuck2019ccm}                   & RGB   & 9.84 & 0.76 & 5.30 & \ding{55} & \ding{55} & \ding{55} & 2.12 & 9.31 & 5.71 & 0.51 & 0.48 & 0.49 \\
MultiSlam-DiffPose~\cite{lipson2024multi} & RGB & \cellcolor{colorTrd}0.41 & 0.75 & 0.58 & 0.68 & \cellcolor{colorSnd}0.25 & \cellcolor{colorTrd}0.47 & \cellcolor{colorTrd}0.63 & \cellcolor{colorTrd}0.68 & \cellcolor{colorTrd}0.65 & \cellcolor{colorSnd}0.32 & \cellcolor{colorTrd}0.40 & \cellcolor{colorTrd}0.36 \\
MA-MASt3R-SLAM~\cite{zhou2026multi}            & RGB   & 3.79 & 1.34 & 2.57 & 3.27 & 7.20 & 5.24 & 4.60 & 6.35 & 5.48 & -- & -- & -- \\
\midrule
\textbf{MAGS-SLAM (Ours)}                             & RGB   & 0.47 & \cellcolor{colorFst}\textbf{0.12} & \cellcolor{colorSnd}0.29 & 0.50 & \cellcolor{colorTrd}0.42 & \cellcolor{colorSnd}0.46 & \cellcolor{colorSnd}0.59 & \cellcolor{colorSnd}0.34 & \cellcolor{colorSnd}0.46 & \cellcolor{colorFst}\textbf{0.29} & \cellcolor{colorSnd}0.34 & \cellcolor{colorFst}\textbf{0.31} \\
\bottomrule
\end{tabular}
\vspace{-1.2\baselineskip}
\end{table*}

\begin{table}[t]
\caption{
    Multi-agent tracking accuracy comparison (ATE RMSE [cm] $\downarrow$) on \underline{real-world} AriaMultiagent datasets. Best results are highlighted as \colorbox{colorFst}{\textbf{first}}, \colorbox{colorSnd}{second}, and \colorbox{colorTrd}{third}.  "--" indicates that CP-SLAM does not support settings with more than two agent. "\ding{55}" indicates invalid results due to the failure.
    }
\vspace{-0.3cm}
\label{tab:tracking_aria}
\setlength{\tabcolsep}{1pt}
\centering
\footnotesize
\begin{tabular}{l c | cccc | cccc }
\toprule
\multirow{2}{*}{\textbf{Method}} &
\multirow{2}{*}{\textbf{Input}} &
\multicolumn{4}{c|}{\textbf{Room-0}} &
\multicolumn{4}{c}{\textbf{Room-1}} \\
& & Ag.1 & Ag.2 & Ag.3 & Avg. & Ag.1 & Ag.2 & Ag.3 & Avg. \\
\midrule
Swarm-SLAM                 & RGB-D & 6.11 & 8.43 & 4.82 & 6.45 & 4.29 & 4.95 & 5.12 & 4.78 \\
CP-SLAM                    & RGB-D & \cellcolor{colorSnd}0.68 & 5.39 & -- & 3.03 & 5.06 & 0.68 & -- & 2.87 \\
MNE-SLAM                   & RGB-D & 1.93 & 2.52 & \cellcolor{colorTrd}0.98 & 1.81 & 1.24 & 0.72 & \cellcolor{colorTrd}0.91 & 0.95 \\
MAGiC-SLAM                 & RGB-D & \cellcolor{colorFst}\textbf{0.67} & \cellcolor{colorSnd}1.13 & 1.67 & \cellcolor{colorTrd}1.15 & \cellcolor{colorTrd}0.96 & \cellcolor{colorSnd}0.53 & \cellcolor{colorSnd}0.46 & \cellcolor{colorSnd}0.65 \\
\midrule
CCM-SLAM                   & RGB   & \ding{55} & \ding{55} & \ding{55} & \ding{55} & \ding{55} & \ding{55} & \ding{55} & \ding{55} \\
MultiSlam-DiffPose         & RGB   & 1.07 & \cellcolor{colorTrd}1.30 & \cellcolor{colorSnd}0.79 & \cellcolor{colorSnd}1.05 & \cellcolor{colorSnd}0.84 & \cellcolor{colorTrd}0.60 & 1.11 & \cellcolor{colorTrd}0.85 \\
\midrule
\textbf{Ours}               & RGB   & \cellcolor{colorTrd}1.02 & \cellcolor{colorFst}\textbf{1.04} & \cellcolor{colorFst}\textbf{0.51} & \cellcolor{colorFst}\textbf{0.86} & \cellcolor{colorFst}\textbf{0.83} & \cellcolor{colorFst}\textbf{0.52} & \cellcolor{colorFst}\textbf{0.26} & \cellcolor{colorFst}\textbf{0.54} \\
\bottomrule
\end{tabular}
\vspace{-1.2\baselineskip}
\end{table}

\begin{table*}[!t]
\caption{
    Multi-agent rendering quality comparison (PSNR/SSIM/LPIPS $\uparrow\uparrow\downarrow$) on ReplicaMultiagent datasets. Best results are highlighted as \colorbox{colorFst}{\textbf{first}}, \colorbox{colorSnd}{second}, and \colorbox{colorTrd}{third}.
    }
\vspace{-0.3cm}
\label{tab:rendering_replica}
\setlength{\tabcolsep}{3pt}
\centering
\footnotesize
\begin{tabular}{l c | ccc | ccc | ccc | ccc}
\toprule
\multirow{2}{*}{\textbf{Method}} &
\multirow{2}{*}{\textbf{Input}} &
\multicolumn{3}{c|}{\textbf{Office-0}} &
\multicolumn{3}{c|}{\textbf{Apartment-0}} &
\multicolumn{3}{c|}{\textbf{Apartment-1}} &
\multicolumn{3}{c}{\textbf{Apartment-2}} \\
& & PSNR $\uparrow$ & SSIM $\uparrow$ & LPIPS $\downarrow$ & PSNR $\uparrow$ & SSIM $\uparrow$ & LPIPS $\downarrow$ & PSNR $\uparrow$ & SSIM $\uparrow$ & LPIPS $\downarrow$ & PSNR $\uparrow$ & SSIM $\uparrow$ & LPIPS $\downarrow$ \\
\midrule
CP-SLAM~\cite{hu2023cp} & RGB-D & 32.61 & 0.90 & 0.29 & 32.16 & \cellcolor{colorTrd}{0.91} & 0.23 & 27.42 & 0.78 & 0.36 & 25.12 & 0.77 & 0.37 \\
MNE-SLAM~\cite{deng2025mne} & RGB-D & 28.27 & 0.86 & 0.08 & 27.46 & 0.87 & 0.27 & 25.45 & 0.73 & 0.41 & 26.16 & \cellcolor{colorTrd}{0.81} & 0.31 \\
MAGiC-SLAM~\cite{yugay2025magic} & RGB-D & 39.32 & \cellcolor{colorSnd}{0.99} & \cellcolor{colorTrd}{0.05} & \cellcolor{colorTrd}{36.96} & \cellcolor{colorSnd}{0.98} & 0.09 & \cellcolor{colorTrd}{30.01} & 0.95 & \cellcolor{colorTrd}{0.18} & 30.73 & \cellcolor{colorSnd}{0.96} & 0.17 \\
MAC-Ego3D~\cite{xu2025mac} & RGB-D & \cellcolor{colorTrd}{42.43} & \cellcolor{colorTrd}{0.98} & \cellcolor{colorSnd}{0.03} & \cellcolor{colorFst}{\textbf{42.83}} & \cellcolor{colorSnd}{0.98} & \cellcolor{colorTrd}{0.05} & \cellcolor{colorFst}{\textbf{36.28}} & \cellcolor{colorSnd}{0.97} & \cellcolor{colorFst}{\textbf{0.06}} & \cellcolor{colorSnd}{38.59} & \cellcolor{colorFst}{\textbf{0.98}} & \cellcolor{colorSnd}{0.06} \\
CoMA-SLAM~\cite{chen2026coma} & RGB-D & \cellcolor{colorSnd}{42.77} & \cellcolor{colorFst}{\textbf{1.00}} & \cellcolor{colorFst}{\textbf{0.02}} & \cellcolor{colorSnd}{42.28} & \cellcolor{colorFst}{\textbf{0.99}} & \cellcolor{colorSnd}{0.03} & \cellcolor{colorSnd}{33.86} & \cellcolor{colorFst}{\textbf{0.98}} & \cellcolor{colorSnd}{0.08} & \cellcolor{colorTrd}{34.59} & \cellcolor{colorFst}{\textbf{0.98}} & \cellcolor{colorTrd}{0.08} \\
\midrule
\textbf{MAGS-SLAM (Ours)} & RGB & \cellcolor{colorFst}{\textbf{43.24}} & \cellcolor{colorSnd}{0.99} & \cellcolor{colorFst}{\textbf{0.02}} & \cellcolor{colorFst}{\textbf{42.83}} & \cellcolor{colorFst}{\textbf{0.99}} & \cellcolor{colorFst}{\textbf{0.02}} & \cellcolor{colorFst}{\textbf{36.28}} & \cellcolor{colorTrd}{0.96} & \cellcolor{colorFst}{\textbf{0.06}} & \cellcolor{colorFst}{\textbf{38.99}} & \cellcolor{colorFst}{\textbf{0.98}} & \cellcolor{colorFst}{\textbf{0.04}} \\
\bottomrule
\end{tabular}
\vspace{-1.2\baselineskip}
\end{table*}

\begin{table}[t]
\caption{
    Multi-agent rendering quality comparison (PSNR/SSIM/LPIPS $\uparrow\uparrow\downarrow$) on \underline{real-world} AriaMultiagent datasets. Best results are highlighted as \colorbox{colorFst}{\textbf{first}}, \colorbox{colorSnd}{second}, and \colorbox{colorTrd}{third}.
    }
\vspace{-0.3cm}
\label{tab:rendering_aria}
\setlength{\tabcolsep}{0.4pt}
\centering
\footnotesize
\begin{tabular}{l c | ccc | ccc }
\toprule
\multirow{2}{*}{\textbf{Method}} &
\multirow{2}{*}{\textbf{Input}} &
\multicolumn{3}{c|}{\textbf{Room-0}} &
\multicolumn{3}{c}{\textbf{Room-1}} \\
& & PSNR $\uparrow$ & SSIM $\uparrow$ & LPIPS $\downarrow$ & PSNR $\uparrow$ & SSIM $\uparrow$ & LPIPS $\downarrow$ \\
\midrule
CP-SLAM & RGB-D & 8.96 & 0.32 & 0.91 & 9.17 & 0.24 & 0.95 \\
MNE-SLAM & RGB-D & 19.23 & 0.78 & 0.39 & \cellcolor{colorTrd}{22.02} & 0.84 & 0.27 \\
MAGiC-SLAM & RGB-D & \cellcolor{colorTrd}{23.45} & \cellcolor{colorTrd}{0.89} & \cellcolor{colorTrd}{0.22} & 21.78 & \cellcolor{colorTrd}{0.85} & \cellcolor{colorTrd}{0.21} \\
MAC-Ego3D & RGB-D & 22.85 & 0.81 & 0.36 & \cellcolor{colorSnd}{28.17} & \cellcolor{colorSnd}{0.91} & \cellcolor{colorSnd}{0.20} \\
CoMA-SLAM & RGB-D & \cellcolor{colorSnd}{25.41} & \cellcolor{colorSnd}{0.90} & \cellcolor{colorSnd}{0.21} & 21.80 & \cellcolor{colorTrd}{0.85} & \cellcolor{colorSnd}{0.20} \\
\midrule
\textbf{Ours} & RGB & \cellcolor{colorFst}{\textbf{29.03}} & \cellcolor{colorFst}{\textbf{0.92}} & \cellcolor{colorFst}{\textbf{0.19}} & \cellcolor{colorFst}{\textbf{31.02}} & \cellcolor{colorFst}{\textbf{0.95}} & \cellcolor{colorFst}{\textbf{0.11}} \\
\bottomrule
\end{tabular}
\vspace{-1.2\baselineskip}
\end{table}

\begin{table}[t]
\caption{
    Multi-agent rendering quality comparison (PSNR/SSIM/LPIPS $\uparrow\uparrow\downarrow$) on \underline{real-world} ETH3D datasets. Best results are highlighted as \colorbox{colorFst}{\textbf{first}}, \colorbox{colorSnd}{second}, and \colorbox{colorTrd}{third}.
    }
\vspace{-0.3cm}
\label{tab:rendering_eth3d}
\setlength{\tabcolsep}{2.2pt}
\centering
\footnotesize
\begin{tabular}{c c | c c c c c}
\toprule
\textbf{Method} & \textbf{Metric} & \textbf{Sofa} & \textbf{Table} & \textbf{Plant-Scene} & \textbf{Mannequin} & \textbf{Planar} \\
\midrule
\multirow{3}{*}{\makecell{CP-SLAM\\RGB-D}}
& PSNR $\uparrow$ & \cellcolor{colorTrd}{25.28} & 15.44 & 20.10 & 18.76 & 16.74 \\
& SSIM $\uparrow$ & \cellcolor{colorTrd}{0.819} & 0.522 & 0.654 & 0.598 & 0.431 \\
& LPIPS $\downarrow$ & \cellcolor{colorTrd}{0.330} & 0.621 & 0.499 & 0.605 & 0.706 \\
\midrule
\multirow{3}{*}{\makecell{MNE-SLAM\\RGB-D}}
& PSNR $\uparrow$ & 22.30 & \cellcolor{colorTrd}{19.48} & 22.72 & \cellcolor{colorTrd}{24.35} & 15.18 \\
& SSIM $\uparrow$ & 0.747 & 0.672 & 0.746 & 0.757 & 0.318 \\
& LPIPS $\downarrow$ & 0.424 & 0.493 & 0.399 & 0.486 & 0.846 \\
\midrule
\multirow{3}{*}{\makecell{MAGiC-SLAM\\RGB-D}}
& PSNR $\uparrow$ & 25.25 & 18.42 & \cellcolor{colorTrd}{24.62} & 23.68 & \cellcolor{colorSnd}{17.35} \\
& SSIM $\uparrow$ & 0.799 & \cellcolor{colorTrd}{0.680} & \cellcolor{colorTrd}{0.849} & \cellcolor{colorTrd}{0.784} & \cellcolor{colorTrd}{0.588} \\
& LPIPS $\downarrow$ & 0.496 & \cellcolor{colorTrd}{0.491} & \cellcolor{colorTrd}{0.352} & \cellcolor{colorTrd}{0.477} & \cellcolor{colorSnd}{0.542} \\
\midrule
\multirow{3}{*}{\makecell{MAC-Ego3D\\RGB-D}}
& PSNR $\uparrow$ & \cellcolor{colorSnd}{27.09} & \cellcolor{colorSnd}{20.46} & \cellcolor{colorSnd}{26.02} & \cellcolor{colorSnd}{27.20} & \cellcolor{colorTrd}{17.04} \\
& SSIM $\uparrow$ & \cellcolor{colorSnd}{0.926} & \cellcolor{colorSnd}{0.851} & \cellcolor{colorSnd}{0.931} & \cellcolor{colorSnd}{0.929} & \cellcolor{colorSnd}{0.621} \\
& LPIPS $\downarrow$ & \cellcolor{colorSnd}{0.178} & \cellcolor{colorSnd}{0.218} & \cellcolor{colorSnd}{0.139} & \cellcolor{colorFst}{\textbf{0.144}} & \cellcolor{colorTrd}{0.569} \\
\midrule
\multirow{3}{*}{\makecell{\textbf{Ours}\\RGB}}
& PSNR $\uparrow$ & \cellcolor{colorFst}{\textbf{30.67}} & \cellcolor{colorFst}{\textbf{27.05}} & \cellcolor{colorFst}{\textbf{31.32}} & \cellcolor{colorFst}{\textbf{31.99}} & \cellcolor{colorFst}{\textbf{22.68}} \\
& SSIM $\uparrow$ & \cellcolor{colorFst}{\textbf{0.953}} & \cellcolor{colorFst}{\textbf{0.938}} & \cellcolor{colorFst}{\textbf{0.961}} & \cellcolor{colorFst}{\textbf{0.939}} & \cellcolor{colorFst}{\textbf{0.864}} \\
& LPIPS $\downarrow$ & \cellcolor{colorFst}{\textbf{0.138}} & \cellcolor{colorFst}{\textbf{0.154}} & \cellcolor{colorFst}{\textbf{0.108}} & \cellcolor{colorSnd}{0.191} & \cellcolor{colorFst}{\textbf{0.222}} \\
\bottomrule
\end{tabular}
\vspace{-1.4\baselineskip}
\end{table}

\subsubsection{MaPGBA consistency}
\label{sec:reactive}
When an agent completes an agent block of Eq.~\eqref{eq:mapgba_bcd} on submap \(\mathcal{P}^a_\ell\), it reports the bulk \(\mathrm{Sim}(3)\) increment that Eq.~\eqref{eq:pgba} applied to that submap's local frame, obtained in closed form by Umeyama~\cite{umeyama1991least} on its pre/post-PGBA keyframe camera centers \(\mathbf{c}^{\text{pre}}_k,\mathbf{c}^{\text{post}}_k\),
\begin{equation}
\Delta_{a,\ell} =
\arg\min_{\Delta\in\mathrm{Sim}(3)}
\frac{1}{|\mathcal{P}^a_\ell|}\sum_{k\in\mathcal{P}^a_\ell}
\bigl\|\mathbf{c}^{\text{post}}_k - \Delta\cdot\mathbf{c}^{\text{pre}}_k\bigr\|^2,
\label{eq:rigidity_fit_main}
\end{equation}
whose square-root optimum \(\rho_{\text{rig}}\) is the rigidity residual, i.e.\ the extent to which the closure fails to act on the submap as a single similarity transform, {and the bulk update is accepted only when \(\rho_{\text{rig}}\le\tau_{\text{rig}}\).
{The threshold \(\tau_{\text{rig}}\) is constrained by the verification cascade that admitted the measurement: once a correspondence is displaced by more than the cascade's inlier threshold, the stored measurement can no longer be reliably transported under a single submap-level similarity transform. Accordingly, with \(\rho_{\text{rig}}\le\tau_{\text{in}}\sqrt{1-r_{\min}}\) (Table~\ref{tab:params}), \(\tau_{\text{rig}}\) is set to \(0.1\,\mathrm{m}\).}
When this rigidity test fails, all incident edges \(e\!\in\!\mathcal{E}^v\) are invalidated and queued for re-verification; otherwise, the accepted bulk shift is used to update each stored \(M_e\) analytically,}
\begin{equation}
M_e \leftarrow
\begin{cases}
M_e \circ \Delta_{a,\ell}^{-1}, & a = \text{src}(e),\\
\Delta_{a,\ell} \circ M_e, & a = \text{tgt}(e).
\end{cases}
\label{eq:edge_update}
\end{equation}
Equation~\eqref{eq:edge_update} is precisely the transport required by MaPGBA consistency: under the rigid hypothesis \(\mathbf{p}\mapsto\Delta_{a,\ell}\cdot\mathbf{p}\), the rewritten measurement encodes the same physical correspondences in the new local frame. Together with the gauge change \(C^a_\ell\!\leftarrow\!C^a_\ell\circ\Delta_{a,\ell}^{-1}\), which leaves Eq.~\eqref{eq:mapgba_lift} invariant, the geometric residual of Eq.~\eqref{eq:geo_residual} remains unchanged when \(\mathcal{P}^a_\ell\) is the target endpoint and changes only by the \(\mathrm{Sim}(3)\) adjoint \(\mathrm{Ad}_{\Delta_{a,\ell}}\) when it is the source. Zero-residual edges therefore remain at zero, while well-fit edges remain well fit, allowing the coordinator block to be warm-started rather than restarted after an agent update. Independently of which branch is taken, the cached anchor-image photometric payload of \(\mathcal{P}^a_\ell\) is discarded, since PGBA may shift its anchor pose and inverse depth even when the bulk drift is sufficiently rigid. The payload is then rebuilt lazily at the next evaluation of \(\mathbf{r}^{\text{pho}}_e\). Algorithm~\ref{alg:coordinator} summarizes one coordinator block together with this consistency step.

\subsection{Occupancy-Aware Asynchronous Fusion}
\label{sec:gs_fusion}
After pose-graph convergence, each submap carries a \(\mathrm{Sim}(3)\) correction \(C^a_\ell\).
Applying this correction to a Gaussian yields
\begin{equation}
\bm{\mu}_i' = s\mathbf{R}\bm{\mu}_i + \mathbf{t},\quad
\bm{\Sigma}_i' = s^2 \mathbf{R}\bm{\Sigma}_i\mathbf{R}^\top,\quad
\mathbf{s}_i' = s\mathbf{s}_i,
\label{eq:gauss_update}
\end{equation}
with unchanged opacity and color.
The scale factor is critical for RGB-only operation. Fusion proceeds asynchronously per target agent using a frustum-aware occupancy grid 
of voxel size \(v\).
For the target map we precompute occupied voxels
\begin{equation}
\mathcal{O}_{\text{occ}} = \bigcup_{G_i \in \mathcal{G}^\text{tgt}}
\bigl\{\kappa(\mathbf{x}) :
\mathbf{x} \in [\bm{\mu}_i - k\mathbf{s}_i,
              \bm{\mu}_i + k\mathbf{s}_i]\bigr\}
\label{eq:envelope}
\end{equation}
and free-space voxels carved by ray-casting every keyframe depth map:
\begin{equation}
\mathcal{O}_{\text{free}} = \bigcup_{\mathbf{u}, k}
\bigl\{\kappa\bigl(\mathbf{c}_k + t(\mathbf{r}_k(\mathbf{u}) -
\mathbf{c}_k)\bigr) : t \in [0,1)\bigr\}.
\label{eq:free_carve}
\end{equation}
A borrowed Gaussian \(G_j\) is retained only if
\begin{equation}
\kappa(\bm{\mu}_j) \notin \mathcal{O}_{\text{occ}}
\quad\text{and}\quad
\kappa(\bm{\mu}_j) \notin \mathcal{O}_{\text{free}}.
\label{eq:dedup}
\end{equation}
The occupied test suppresses duplicate Gaussians, while the free-space test prevents
inserting source primitives into regions already observed as empty by the target agent.
We pack each voxel index $\kappa(\cdot)$ into a 64-bit spatial-hash key, reducing the membership tests in Eq.~\eqref{eq:dedup} to constant-time lookups.
After deduplication, small residual \(\mathrm{Sim}(3)\) errors and exposure differences
are resolved by joint photometric refinement of the fused Gaussian set
\(\mathcal{G}^{\text{fused}}=\mathcal{G}^{\text{tgt}}\cup\{G_j:\text{Eq.~\eqref{eq:dedup}}\}\),
per-keyframe pose deltas \(\delta_{\xi_k}\), and per-keyframe log-gain and bias \((a_k,b_k)\in\mathbb{R}^2\), with
\(\tilde{I}(\mathbf{u}) = e^{a_k}\hat{I}(\mathbf{u}) + b_k\) and \(\hat{I},\hat{D}\) the differentiably rendered RGB and depth of \(\mathcal{G}^{\text{fused}}\).
The refinement loss reuses the photometric, depth, and normal terms of Eq.~\eqref{eq:map_loss} but evaluates them on the exposure-compensated render \(\tilde{I}\),
and adds a structural-similarity term:
\begin{equation}
\mathcal{L}^{\text{rf}} =
\alpha'\mathcal{L}_1(\tilde{I}, I)
+ \lambda\bigl(1-\mathrm{SSIM}(\tilde{I}, I)\bigr)
+ \beta\mathcal{L}_1(\hat{D}, \bar{D})
+ \lambda_n'\mathcal{L}_{\text{normal}},
\label{eq:refine_loss}
\end{equation}
with \(\alpha' = (1-\lambda)+\alpha = 1.75\), \(\lambda=0.2\), \(\beta\) inherited from Eq.~\eqref{eq:map_loss},
and {\(\lambda_n' = 10\lambda_n = 0.1\)} raised to anchor geometry against photometric drift. \(\bar{D}\) is derived from the scaled \(D\).
The refinement runs for 2{,}000 iterations at half the per-agent learning rate and prunes Gaussians
with opacity below \(0.005\), producing a globally consistent map.
\section{Analysis}
\label{sec:experiments}

{
\subsection{Evaluation Protocol}
\label{sec:protocol}

\subsubsection{Depth for RGB-D baselines}
On ReplicaMultiagent and ReplicaMultiagent Plus, the RGB-D baselines use the noise-free metric depth provided by the simulator, without an additional sensor-noise model. For the real-world benchmarks, the RGB-D baselines use the benchmark-provided metric depth: sensor-measured depth for ETH3D and ground-truth depth generated by the Aria Digital Twin system for AriaMultiagent.

\subsubsection{Rendering evaluation}
PSNR (Peak Signal-to-Noise Ratio), SSIM (Structural Similarity Index), and LPIPS (Learned Perceptual Image Patch Similarity) are evaluated on the recorded image stream of each agent using the estimated camera poses. We report the average across all evaluated frames and agents, following the protocol of MAGiC-SLAM~\cite{yugay2025magic}.

\subsubsection{Geometry evaluation}
Accuracy, completeness, and normal consistency are computed from meshes reconstructed through a common evaluation pipeline. For each method, the depth maps rendered at the evaluated views are TSDF-fused with a \(2\,\mathrm{cm}\) voxel size, and the resulting mesh is transformed into the ground-truth frame using a dedicated \(\mathrm{Sim}(3)\) alignment estimated with Umeyama~\cite{umeyama1991least}. Accuracy is the mean distance from \(2\times10^{5}\) points sampled on the reconstructed mesh to the ground-truth surface, completeness is the mean distance in the opposite direction, and normal consistency is the bidirectional mean absolute cosine similarity between corresponding surface normals.
}

\subsection{Tracking Accuracy and Localization}
As the first RGB-only multi-agent 3D Gaussian Splatting SLAM system based on monocular input, MAGS-SLAM achieves highly competitive trajectory estimation using RGB observations alone.
Table~\ref{tab:tracking_replica} reports the Absolute Trajectory Error (ATE) RMSE [cm] on the ReplicaMultiagent benchmark, {following the same evaluation protocol as MAGiC-SLAM~\cite{yugay2025magic}.}
{Without depth sensors,} our method matches or outperforms several RGB-D baselines and surpasses previous RGB-only approaches, such as CCM-SLAM~\cite{schmuck2019ccm}, MultiSLAM-DiffPose~\cite{lipson2024multi}, and MA-MASt3R-SLAM~\cite{zhou2026multi}.
Similar trends hold on the real-world AriaMultiagent dataset (Table~\ref{tab:tracking_aria}), where MAGS-SLAM maintains robust multi-agent alignment even {in challenging real-world indoor environments}.
Notably, on Room-0 and Room-1 sequences, MAGS-SLAM attains average errors of 0.86\,cm and 0.54\,cm, respectively, which are on par with or better than {those of} the RGB-D state-of-the-art methods MNE-SLAM~\cite{deng2025mne} and MAGiC-SLAM~\cite{yugay2025magic}.

\subsection{Photorealistic Mapping and Rendering Quality}

Tables~\ref{tab:rendering_replica}, \ref{tab:rendering_aria}, and \ref{tab:rendering_eth3d} report rendering quality on ReplicaMultiagent, AriaMultiagent, and ETH3D. Despite using only RGB input, MAGS-SLAM consistently ranks among the top-performing PSNR, SSIM, and LPIPS across all three benchmarks. On ReplicaMultiagent it reaches 43.24/42.83/38.99\,dB PSNR on Office-0, Apartment-0, and Apartment-2, matching or surpassing RGB-D baselines such as MAGiC-SLAM~\cite{yugay2025magic}, MAC-Ego3D~\cite{xu2025mac}, and CoMA-SLAM~\cite{chen2026coma}. On the real-world AriaMultiagent sequences, it achieves 29.03 and 31.02\,dB on Room-0 and Room-1, exceeding the strongest RGB-D competitor by 3.6 and 2.9\,dB. On the more challenging ETH3D multi-agent benchmark it ranks first on all five scenes, with 3-7\,dB PSNR gains over MAC-Ego3D. Qualitative comparisons (Fig.~\ref{fig:replica_results} and \ref{fig:eth3d_aria_results}) confirm these gains: our occupancy-aware fusion and joint pose-Gaussian refinement yield seamless reconstructions free of ghosting seams, while MAGiC-SLAM and MAC-Ego3D show misalignment and duplicated Gaussians at agent boundaries that the frustum-aware occupancy grid and final refinement together eliminate.

\subsection{Scaling to Four Agents on ReplicaMultiagent Plus}
\label{sec:replica_plus}
\begin{figure}[t]
\centering
\includegraphics[width=0.47\textwidth]{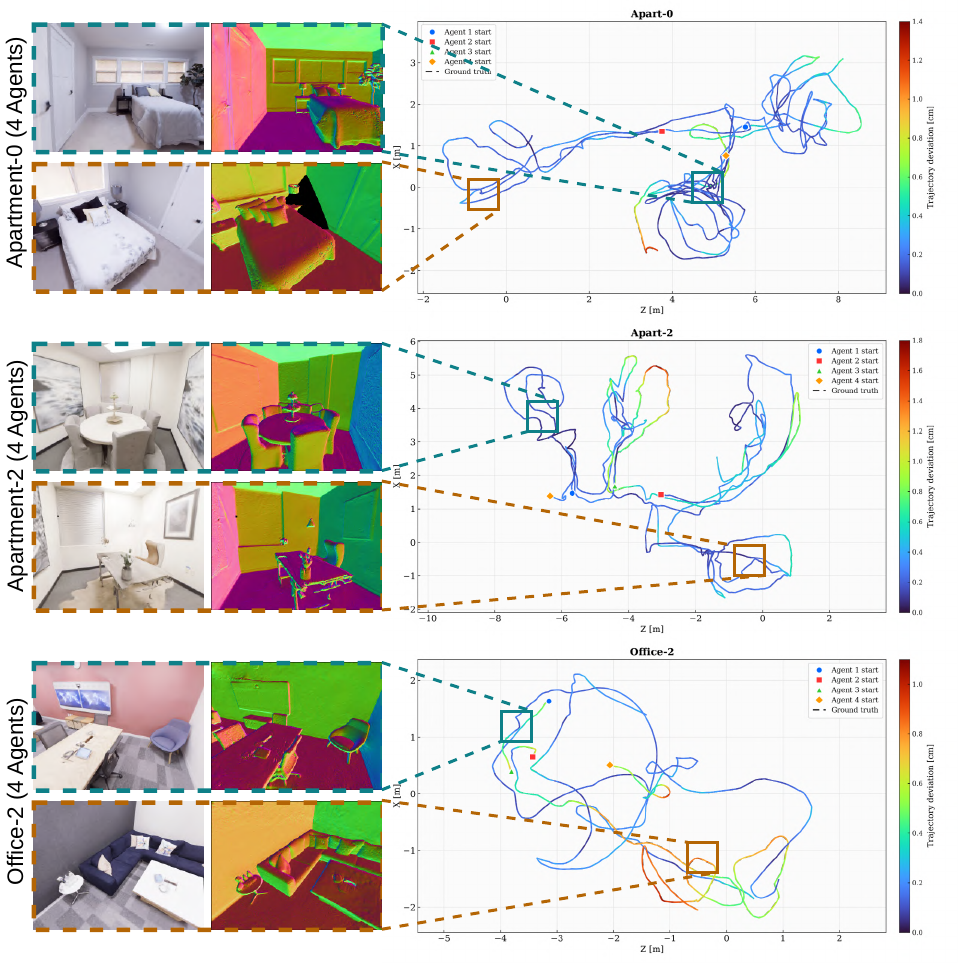}
\vspace{-0.8\baselineskip}
\caption{\textbf{Tracking Performance on the proposed ReplicaMultiagent Plus benchmark.}
Top-down views of the four estimated agent trajectories for each scene, coloured by per-frame trajectory deviation from ground truth.}
\label{fig:replica_plus_comparison}
\vspace{-1.2\baselineskip}
\end{figure}

To evaluate robustness as the number of collaborating agents grows, we test on our ReplicaMultiagent Plus benchmark, which extends Replica to four agents with long trajectories and dense ground-truth geometry and semantics; its construction is detailed in Appendix~\ref{sec:supp_dataset}. Table~\ref{tab:rendering_replica_plus} reports tracking (ATE) and rendering (PSNR/SSIM/LPIPS) jointly on three four-agent scenes, and Fig.~\ref{fig:replica_plus_comparison} shows the corresponding qualitative results. {Despite relying on four independently scaled monocular submaps without depth sensors}, MAGS-SLAM attains the best rendering quality on all three scenes, reaching 40.19/40.86/41.07\,dB PSNR on Office-2, Apartment-0, and Apartment-2 and exceeding the strongest RGB-D competitor. On trajectory accuracy it stays competitive with the RGB-D baselines, obtaining the lowest ATE on Office-2 (0.354\,cm) and Apartment-0 (0.282\,cm) and a close third on Apartment-2. The RGB baselines MultiSLAM-DiffPose provide no renderable map.

{
\subsection{Geometric Reconstruction Quality}

Table~\ref{tab:geometry} reports geometric reconstruction quality on ReplicaMultiagent Plus in terms of accuracy, completeness, and normal consistency. Despite relying solely on RGB input, MAGS-SLAM achieves the best or competitive average performance across all three metrics, with an accuracy of \(2.96\,\mathrm{cm}\), completeness of \(2.84\,\mathrm{cm}\), and normal consistency of \(95.31\%\). In particular, it attains the best completeness and normal consistency on Office-2, reaching \(2.68\,\mathrm{cm}\) and \(95.58\%\), respectively, while remaining competitive in accuracy. On Apartment-0, MAGS-SLAM achieves \(3.63\,\mathrm{cm}\) accuracy, \(3.03\,\mathrm{cm}\) completeness, and \(94.90\%\) normal consistency, closely matching or surpassing the strongest RGB-D baseline, MAGiC-SLAM~\cite{yugay2025magic}. On Apartment-2, it further obtains the best results across all three metrics, with \(2.28\,\mathrm{cm}\) accuracy, \(2.80\,\mathrm{cm}\) completeness, and \(95.45\%\) normal consistency. These results demonstrate that, even without depth sensors, MAGS-SLAM reconstructs geometrically accurate, complete, and locally consistent surfaces comparable to or better than state-of-the-art RGB-D collaborative SLAM systems.
}

\begin{table*}[!t]
\caption{
    Multi-agent tracking accuracy comparison (ATE RMSE [cm] $\downarrow$) and rendering quality comparison (PSNR/SSIM/LPIPS $\uparrow\uparrow\downarrow$) on ReplicaMultiagent Plus datasets. Best results are highlighted as \colorbox{colorFst}{\textbf{first}}, \colorbox{colorSnd}{second}, and \colorbox{colorTrd}{third}. "--" indicates that MultiSlam-DiffPose lack rendering capabilities. "$\S$" indicates CP-SLAM does not support settings with more than two agent.
    }
\vspace{-0.3cm}
\label{tab:rendering_replica_plus}
\setlength{\tabcolsep}{1.0pt}
\centering
\small
\begin{tabular}{l c | cccc | cccc | cccc}
\toprule
\multirow{2}{*}{\textbf{Method}} &
\multirow{2}{*}{\textbf{Input}} &
\multicolumn{4}{c|}{\textbf{Office-2 (4 Agents)}} &
\multicolumn{4}{c|}{\textbf{Apartment-0 (4 Agents)}} &
\multicolumn{4}{c}{\textbf{Apartment-2 (4 Agents)}} \\
& & ATE $\downarrow$ & PSNR $\uparrow$ & SSIM $\uparrow$ & LPIPS $\downarrow$
& ATE $\downarrow$ & PSNR $\uparrow$ & SSIM $\uparrow$ & LPIPS $\downarrow$
& ATE $\downarrow$ & PSNR $\uparrow$ & SSIM $\uparrow$ & LPIPS $\downarrow$ \\
\midrule
CP-SLAM$^\S$~\cite{hu2023cp} & RGB-D
& 2.591 & 20.44 & 0.763 & 0.391 & \cellcolor{colorSnd}0.463 & \cellcolor{colorSnd}29.19 & \cellcolor{colorTrd}0.918 & \cellcolor{colorSnd}0.165 & 1.098 & 22.89 & \cellcolor{colorTrd}0.818 & 0.368 \\
MNE-SLAM~\cite{deng2025mne} & RGB-D
& 0.508 & 27.71 & 0.907 & 0.203
& 0.974 & 25.46 & 0.883 & 0.218
& 0.614 & \cellcolor{colorTrd}23.27 & 0.756 & 0.402 \\
MAGiC-SLAM~\cite{yugay2025magic} & RGB-D
& \cellcolor{colorSnd}0.395 & \cellcolor{colorTrd}31.37 & \cellcolor{colorSnd}0.969 & \cellcolor{colorTrd}0.101
& 0.612 & \cellcolor{colorTrd}27.09 & 0.909 & \cellcolor{colorTrd}0.188
& \cellcolor{colorSnd}0.365 & 21.04 & 0.762 & \cellcolor{colorTrd}0.366 \\
MAC-Ego3D~\cite{xu2025mac} & RGB-D
& \cellcolor{colorTrd}0.485 & \cellcolor{colorSnd}32.17 & \cellcolor{colorTrd}0.955 & \cellcolor{colorSnd}0.071
& \cellcolor{colorTrd}0.594 & 26.53 & \cellcolor{colorSnd}0.929 & 0.198
& \cellcolor{colorFst}\textbf{0.329} & \cellcolor{colorSnd}27.50 & \cellcolor{colorSnd}0.935 & \cellcolor{colorSnd}0.191 \\
\midrule
MultiSlam-DiffPose \cite{lipson2024multi} & RGB
& 0.892 & -- & -- & -- & 0.905 & -- & -- & -- & 1.023 & -- & -- & -- \\
\midrule
\textbf{MAGS-SLAM (Ours)} & RGB
& \cellcolor{colorFst}\textbf{0.354} & \cellcolor{colorFst}\textbf{40.19} & \cellcolor{colorFst}\textbf{0.988} & \cellcolor{colorFst}\textbf{0.028}
& \cellcolor{colorFst}\textbf{0.282} & \cellcolor{colorFst}\textbf{40.86} & \cellcolor{colorFst}\textbf{0.992} & \cellcolor{colorFst}\textbf{0.013}
& \cellcolor{colorTrd}0.404 & \cellcolor{colorFst}\textbf{41.07} & \cellcolor{colorFst}\textbf{0.990} & \cellcolor{colorFst}\textbf{0.018} \\
\bottomrule
\end{tabular}
\vspace{-1.2\baselineskip}
\end{table*}

\begin{table*}[t]
\caption{
    {Geometric reconstruction quality (accuracy, completeness (cm), and normal consistency (NC, \%)) on ReplicaMultiagent Plus datasets. Best results are highlighted as \cellcolor{colorFst}{\textbf{first}}, \cellcolor{colorSnd}{second}, and \cellcolor{colorTrd}{third}. "$\S$" indicates CP-SLAM does not support settings with more than two agent.}
    }
\vspace{-0.3cm}
\label{tab:geometry}
\setlength{\tabcolsep}{4.6pt}
\centering
\footnotesize
\begin{tabular}{l c | c c c | c c c | c c c | c c c}
\toprule
& & \multicolumn{3}{c|}{{Office-2}} 
& \multicolumn{3}{c|}{{Apartment-0}} 
& \multicolumn{3}{c|}{{Apartment-2}} 
& \multicolumn{3}{c}{{Average}} \\
{Method} & {Input} 
& {Acc.\,$\downarrow$} & {Comp.\,$\downarrow$} & {NC\,$\uparrow$}
& {Acc.\,$\downarrow$} & {Comp.\,$\downarrow$} & {NC\,$\uparrow$}
& {Acc.\,$\downarrow$} & {Comp.\,$\downarrow$} & {NC\,$\uparrow$}
& {Acc.\,$\downarrow$} & {Comp.\,$\downarrow$} & {NC\,$\uparrow$} \\
\midrule

{CP-SLAM$^\S$~\cite{hu2023cp}} & {RGB-D} 
& {25.59} & {22.17} & {68.63} 
& \cellcolor{colorTrd}{{4.40}} & {4.73} & {84.51} 
& {9.92} & {8.65} & {73.22} 
& {13.30} & {11.85} & {75.45} \\

{MNE-SLAM~\cite{deng2025mne}} & {RGB-D} 
& {4.13} & {3.85} & {89.17} 
& {5.42} & {5.91} & {82.54} 
& {7.35} & {6.29} & {81.58} 
& {5.63} & {5.35} & {84.43} \\

{MAGiC-SLAM~\cite{yugay2025magic}} & {RGB-D} 
& \cellcolor{colorFst}{\textbf{{2.89}}} 
& \cellcolor{colorSnd}{{2.97}} 
& \cellcolor{colorSnd}{{95.36}} 
& \cellcolor{colorFst}{\textbf{{3.55}}} 
& \cellcolor{colorSnd}{{3.53}} 
& \cellcolor{colorSnd}{{94.82}} 
& \cellcolor{colorTrd}{{3.84}} 
& \cellcolor{colorSnd}{{2.82}} 
& \cellcolor{colorTrd}{{93.85}} 
& \cellcolor{colorSnd}{{3.42}} 
& \cellcolor{colorSnd}{{3.11}} 
& \cellcolor{colorSnd}{{94.68}} \\

{MAC-Ego3D~\cite{xu2025mac}} & {RGB-D} 
& \cellcolor{colorSnd}{{2.91}} 
& \cellcolor{colorTrd}{{3.20}} 
& \cellcolor{colorTrd}{{93.69}} 
& {6.60} 
& \cellcolor{colorTrd}{{4.23}} 
& \cellcolor{colorTrd}{{86.22}} 
& \cellcolor{colorSnd}{{2.35}} 
& \cellcolor{colorTrd}{{3.66}} 
& \cellcolor{colorSnd}{{94.50}} 
& \cellcolor{colorTrd}{{3.95}} 
& \cellcolor{colorTrd}{{3.70}} 
& \cellcolor{colorTrd}{{91.47}} \\

\midrule

{\textbf{MAGS-SLAM (Ours)}} & {\textbf{RGB}} 
& \cellcolor{colorTrd}{{2.98}} 
& \cellcolor{colorFst}{\textbf{{2.68}}} 
& \cellcolor{colorFst}{\textbf{{95.58}}} 
& \cellcolor{colorSnd}{{3.63}} 
& \cellcolor{colorFst}{\textbf{{3.03}}} 
& \cellcolor{colorFst}{\textbf{{94.90}}} 
& \cellcolor{colorFst}{\textbf{{2.28}}} 
& \cellcolor{colorFst}{\textbf{{2.80}}} 
& \cellcolor{colorFst}{\textbf{{95.45}}} 
& \cellcolor{colorFst}{\textbf{{2.96}}} 
& \cellcolor{colorFst}{\textbf{{2.84}}} 
& \cellcolor{colorFst}{\textbf{{95.31}}} \\

\bottomrule
\end{tabular}
\vspace{-1.0\baselineskip}
\end{table*}

\begin{table*}[t]
\caption{
    Agent-count scaling on ReplicaMultiagent Plus datasets. Per-agent tracking throughput and peak GPU memory, per-candidate verification latency on the coordinator, and per-agent communication overhead as the number of collaborating agents grows from two to four.
}
\vspace{-0.3cm}
\label{tab:scalability}
\setlength{\tabcolsep}{6pt}
\centering
\footnotesize
\begin{tabular}{c | c c c c c}
\toprule
\textbf{Agents} &
\textbf{\shortstack{Track FPS}} &
\textbf{\shortstack{Peak GPU memory (MB)}} &
\textbf{\shortstack{Verification (ms)}} &
\textbf{\shortstack{Coordination uplink (MB)}} &
\textbf{\shortstack{Map-merge payload (MB)}} \\
\midrule
2 & 14.5 & 3833 & 93.6 & 9.27 & 409.8 \\
3 & 14.1 & 3486 & 63.0 & 8.12 & 378.7 \\
4 & 14.5 & 3377 & 53.4 & 7.31 & 265.8 \\
\bottomrule
\end{tabular}
\vspace{-1.2\baselineskip}
\end{table*}

\begin{figure}[t]
    \centering
    \includegraphics[width=\columnwidth]{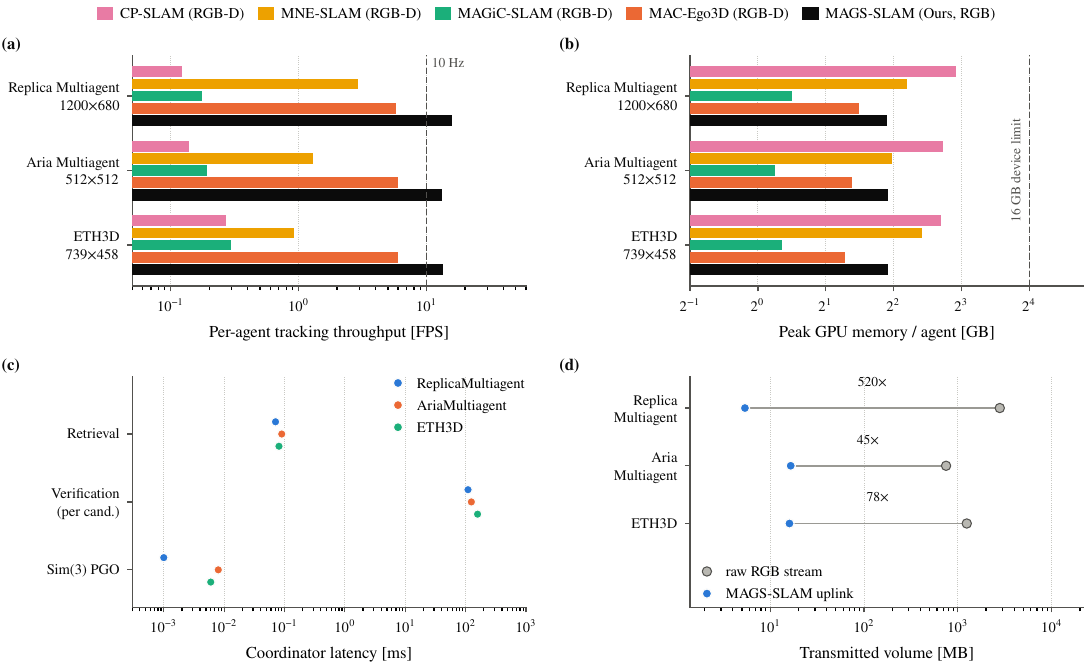}
    \caption{\textbf{Runtime, memory, and communication efficiency of MAGS-SLAM.}
    (a) Per-agent tracking rate stays above the $10$\,Hz real-time line on all three datasets.
    (b) Peak GPU memory per agent is below $3.9$\,GB, far under the $16$\,GB budget of a commodity card, so several agents fit on one device.
    (c) Coordinator cost per operation: Retrieval, per-candidate verification, and full $\mathrm{Sim}(3)$ pose-graph optimization (PGO) are executed asynchronously by the coordinator and are triggered only when new submap candidates arise. Therefore, these operations do not lie on the per-agent tracking critical path and are not included in the reported tracking FPS. Their cost is instead reported separately on a per-operation basis.
    (d) Communication volume: transmitting compact submap summaries instead of the raw RGB stream reduces each agent's uplink by $45$--$520\times$.}
    \label{fig:efficiency}
\vspace{-1.2\baselineskip}
\end{figure}

\subsection{Ablation Study}
\label{sec:ablation_study}
Table~\ref{tab:ablation} ablates each design component on ReplicaMultiagent by disabling one at a time; {each variant degrades either trajectory accuracy, rendering quality, or both.} Removing the Sim(3)-PGBA update, i.e.\ the MaPGBA consistency step of Eq.~\eqref{eq:edge_update} that keeps the two blocks of Eq.~\eqref{eq:mapgba_bcd} expressed in a common local frame, {increases ATE from $0.382$ to $0.398$\,cm and} costs $4.0$\,dB, as residual intra-submap scale drift is left for the rasteriser to absorb. Replacing the photometric inter-agent residual with a purely geometric one {increases ATE to $0.393$\,cm and} drops PSNR to $37.88$\,dB: the photometric term tightens the relative Sim(3) on textured but geometrically ambiguous overlaps such as wall-floor seams, {benefiting both alignment and rendering.} Disabling the frustum-aware occupancy grid in favour of naive Gaussian-centre deduplication is the second-largest hit, since slightly mis-aligned per-agent populations form double-shells on shared surfaces. Skipping the joint pose-Gaussian refinement costs only $1.4$\,dB but supplies the sub-voxel correction needed for sharp rendering. {Its ATE changes only marginally to $0.386$\,cm, further suggesting that this step mainly refines the final representation.} Finally, replacing the Sim(3) graph with a rigid SE(3) one {raises ATE sharply to $4.491$\,cm,} collapses PSNR to $18.61$\,dB and inflates LPIPS by an order of magnitude, confirming that explicit scale estimation is mandatory in the multi-agent monocular regime.

\begin{table}[t]
\caption{
    \textbf{Ablation Studies on ReplicaMultiagent Dataset}. {We ablate the major components of MAGS-SLAM and report their effects on tracking and rendering performance.} Details about each configuration are described in \S\ref{sec:ablation_study}.
    }
\label{tab:ablation}
\setlength{\tabcolsep}{2pt}
\centering
\footnotesize
\resizebox{\columnwidth}{!}{%
\begin{tabular}{l c c c c }
\toprule
Method & {ATE [cm] $\downarrow$} & PSNR $\uparrow$ & SSIM $\uparrow$ & LPIPS $\downarrow$ \\
\midrule
A1. w/o sim(3)-PGBA update & {0.398} & 36.32 & 0.963 & 0.057 \\
A2. w/o photometric residuals & {0.393} & 37.88 & 0.967 & 0.045 \\
A3. w/o occupancy-aware fusion & {0.382} & 33.16 & 0.941 & 0.113 \\
A4. w/o pose-Gaussian joint refinement & {0.386} & 38.94 & 0.974 & 0.041 \\
A5. w/o scale estimation, i.e., SE(3) & {4.491} & 18.61 & 0.718 & 0.390 \\
\midrule
\textbf{Full} & {0.382} & 40.33 & 0.981 & 0.035 \\
\bottomrule
\end{tabular}}
\vspace{-1.2\baselineskip}
\end{table}

\subsection{Real-World Deployment}
\label{sec:real_world}
\begin{figure*}[!t]
    \centering
    \includegraphics[width=\textwidth]{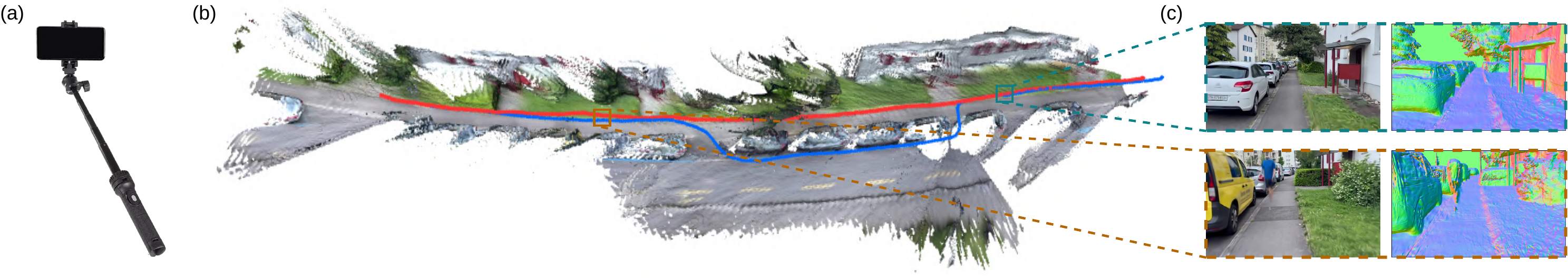}
    \vspace{-1.8\baselineskip}
    \caption{\textbf{Real-world multi-agent reconstruction from casual smartphone capture.}
    Two agents each record a walking sequence of the same outdoor residential
    street using an ordinary smartphone mounted on a handheld monopod~(a),
    providing purely monocular RGB video with no depth sensor or stereo rig. 
    From these two independently captured streams,
    MAGS-SLAM recovers both per-agent camera trajectories (red and blue) and a
    single globally consistent mesh map~(b) that fuses the two traversals
    into one coherent scene. Zooming into two viewpoints~(c), the rendered
    color images (left) reproduce fine appearance detail, parked cars,
    foliage, and building facades, while the rendered surface normals (right)
    reveal clean, floater-free geometry with sharp object boundaries,
    confirming that our occupancy-aware fusion and joint pose-Gaussian
    refinement transfer directly to uncontrolled in-the-wild conditions.}
    \label{fig:real_world}
\end{figure*}

To assess generalization beyond curated benchmarks, we collect a real-world sequence in an uncontrolled outdoor environment using only a consumer smartphone mounted on a handheld monopod (Fig.~\ref{fig:real_world}a). Two agents independently walk along the same residential street, each recording a purely monocular RGB video without any depth sensor, or stereo rig. This in-the-wild setting is substantially harder than our indoor benchmarks: it exhibits outdoor-scale depth, natural illumination changes, repetitive facades, and dynamic elements such as parked cars and a passing pedestrian. Despite operating on RGB alone, MAGS-SLAM aligns the two independently scaled monocular submaps into a single globally consistent Gaussian map and recovers both agent trajectories (Fig.~\ref{fig:real_world}b). The rendered views (Fig.~\ref{fig:real_world}c, left) preserve fine appearance detail, while the rendered surface normals (right) exhibit clean, floater-free geometry with sharp object boundaries, confirming that our frustum-aware occupancy fusion and joint pose-Gaussian refinement transfer directly from synthetic and indoor benchmarks to casually captured, outdoor scenes.

\subsection{Efficiency and Scalability}
\label{sec:efficiency}
Because each agent must run onboard and share a communication link, we profile per-agent runtime, memory, and bandwidth (Fig.~\ref{fig:efficiency}) and study how these scale with the number of agents (Table~\ref{tab:scalability}). As shown in Fig.~\ref{fig:efficiency}, MAGS-SLAM tracks at $13$-$16$\,FPS per agent, above the $10$\,Hz real-time threshold, while keeping peak GPU memory below $3.9$\,GB per agent, so multiple agents comfortably share a single commodity GPU. On the coordinator side, retrieval, per-candidate verification, and full $\mathrm{Sim}(3)$ pose-graph optimization are executed asynchronously and only when new submap candidates arise. Their latency is therefore reported separately from the per-agent tracking FPS; in practice, these operations complete sufficiently quickly that inter-agent collaboration does not stall the front-ends. Crucially, transmitting compact submap summaries rather than raw observations lowers each agent's uplink by $45$-$520\times$, which is what makes the collaboration bandwidth-practical.

Table~\ref{tab:scalability} varies the agent count from two to four on ReplicaMultiagent Plus. Per-agent tracking throughput ($\sim\!14.5$\,FPS) and peak GPU memory ($\sim\!3.4$-$3.8$\,GB) stay essentially flat, and the per-candidate cost does not grow with scale: per-candidate verification even drops from $93.6$ to $53.4$\,ms. {The variation mainly reflects differences in the candidate pairs introduced by each trajectory subset, whose overlap and matching difficulty affect the verification cost.} The coordination uplink and map-merge payload likewise decrease per agent, further demonstrating the favorable scalability of MAGS-SLAM to larger agent teams.

\section{Conclusion}
\label{sec:conclusion}
We propose MAGS-SLAM, the first monocular RGB-only multi-agent 3D Gaussian Splatting SLAM system. By combining a monocular front-end with compact submap summaries, a MaPGBA-consistent Sim(3) submap pose graph augmented by geometric and photometric residuals and a spatial-extent gate, adaptive edge invalidation, and an occupancy-aware asynchronous fusion framework, our method achieves globally consistent collaborative reconstruction from purely monocular video. On ReplicaMultiagent, AriaMultiagent, ETH3D, and our ReplicaMultiagent Plus, {MAGS-SLAM achieves competitive tracking accuracy,
rendering quality, and geometric reconstruction quality compared with} state-of-the-art RGB-D collaborative Gaussian SLAM methods without requiring any depth sensors.
Although MAGS-SLAM already enables practical photorealistic collaboration with ubiquitous RGB cameras, scaling the approach to ultra-large environments remains an open frontier. Future work will explore more efficient global map management, and tighter integration with large foundation models for even more robust scale recovery and semantic understanding, unlocking transformative applications in virtual production and large-scale digital twins.


\appendices
\begin{figure*}[t]
    \centering
    \includegraphics[width=\textwidth]{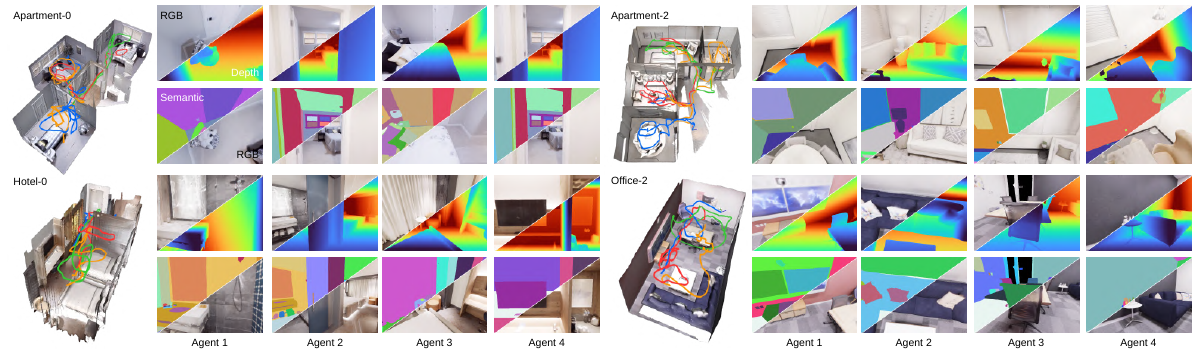}
    \vspace{-1.8\baselineskip}
    \caption{\textbf{Four example scenes from the ReplicaMultiagent Plus benchmark.}
    \emph{Top:} for each scene (Apartment-0, Apartment-2, Office-2, and
    Hotel-0) we show the dense ground-truth mesh with the four
    {human-teleoperated} agent trajectories overlaid, one colour per
    agent. \emph{Bottom:} sample per-agent observations arranged as four
    agent rows $\times$ four scene columns; each tile is split diagonally to
    show the co-registered RGB image, metric depth, and per-pixel semantic
    label exported for every keyframe. {The inter-agent overlap of each
    scene is set at recording time by the choice of start poses and a
    per-scene overlap budget followed by the operator, ranging from largely
    disjoint to heavily co-visible multi-agent splits.}}
    \label{fig:replica_plus_dataset}
    \vspace{-1.2\baselineskip}
\end{figure*}

\section{The ReplicaMultiagent Plus Benchmark}
\label{sec:supp_dataset}

Existing multi-agent SLAM benchmarks force a trade-off between photometric
realism, agent count, and ground-truth completeness. Real-world sets such as
AriaMultiagent~\cite{yugay2025magic, pan2023aria}, the multi-agent split of
ETH3D~\cite{schops2019bad}, and the MNE/MCN-SLAM
dataset~\cite{deng2025mne, deng2025mcn} provide authentic sensor noise but no dense
mesh ground truth, and no semantic supervision. Synthetic sets built on
Replica~\cite{hu2023cp, straub2019replica} provide complete geometry but cap the agent
count at two. Microsoft 7-Scenes~\cite{glocker2013real},
adopted as a multi-agent benchmark by treating its repeated recordings as
independent agents, contributes real-world variability but ships only
per-frame depth without a globally consistent mesh.

We address these gaps by re-rendering Replica scenes through the Habitat-Sim
platform with up to four {human-teleoperated} agent trajectories per
scene. This produces a benchmark that simultaneously exposes (i)~Replica's
native semantic labels, (ii)~dense ground-truth depth and mesh from
the simulator, (iii)~{controlled} trajectory overlap between agents, and
(iv)~more than two {agents per scene}, while preserving the photometric
fidelity. It will support the evaluation of monocular, RGB-D, and semantic multi-agent SLAM for collaborative dense reconstruction.

\subsection{Construction}
\label{sec:supp_dataset_construction}

\subsubsection{Trajectory generation.}
For 4 Replica scenes we instantiate $N_a\in\{2,3,4\}$
embodied agents inside Habitat-Sim~\cite{puig2024habitat}. Their start poses are
drawn from the simulator's navigation mesh and then manually
filtered so that the $N_a$ seeds fall in distinct rooms or hallway
segments. Rather than a purely
procedural random walk, each trajectory is captured by a human operator
teleoperating the agent at $30\mathrm{Hz}$ with an Xbox controller. The operator is asked to cover the navigable area while
{carefully following} a target overlap budget per scene{; together with
the start-pose selection, this sets the inter-agent overlap of each scene,
from largely disjoint to heavily co-visible. The resulting teleoperation}
yields
motion that mixes long forward sweeps, lookarounds, brief stops, and
occasional revisits of the same region. The full $(t, \mathbf{p},
\mathbf{q})$ pose is logged in TUM format alongside the frames, so the
same trajectory can be deterministically re-rendered later at a
different resolution or sensor set without re-recording. {Each of the \(N_a\) trajectories of a scene is teleoperated in its own
episode of the same static scene and logged in that scene's single world
frame.}

\subsubsection{Rendered modalities.}
Each agent carries a rigidly co-located RGB, depth, and semantic camera
with an HFOV of $90^{\circ}$, and frames are rendered at
$640\times480$ at $30\mathrm{Hz}$. For every keyframe we export an
8-bit RGB image, dense metric depth stored as a 16-bit PNG in
millimetres, surface normals, the per-pixel Replica semantic class id,
and the camera pose in a common world frame. Pinhole intrinsics
$(f_x, f_y, c_x, c_y)$ are derived from the configured HFOV and shipped
once per agent, so all $N_a$ agents share the same camera model. {Every agent poses are released in TUM format in Habitat's OpenGL
camera convention. Sharing one camera model keeps the team-size study free of
confounds, but it also means that heterogeneous intrinsics, heterogeneous
fields of view, and rolling-shutter effects are not covered by the
current release.}

\begin{table*}[t]
\caption{{\textbf{ReplicaMultiagent Plus statistics.} Per-agent ranges are
given as min-max over the four agents of the scene. Overlap statistics are computed from the
ground-truth depth at a \(10\,\)cm voxel resolution. Co-visible voxels denotes the
fraction of all observed voxels that are observed by at least two agents; the mean pairwise
overlap averages the voxel overlap of each agent pair over all six pairs.}}
\label{tab:benchmark_stats}
\setlength{\tabcolsep}{8pt}
\centering
\footnotesize
\begin{tabular}{@{}l cccc@{}}
\toprule
{\textbf{Property}} & {\textbf{Office-2}} & {\textbf{Apartment-0}} & {\textbf{Apartment-2}} & {\textbf{Hotel-0}} \\
\midrule
{Agents}                       & {4} & {4} & {4} & {4} \\
{Frames per agent}             & {881--1{,}303} & {1{,}696--2{,}238} & {2{,}062--3{,}049} & {782--1{,}173} \\
{Path length (total) [m]}             & {54.1} & {101.6} & {119.9} & {49.8} \\
{Path per agent [m]}           & {10.9--16.7} & {20.3--30.4} & {25.6--39.2} & {10.4--15.4} \\
{Co-visible voxels [\%]}       & {72.2} & {80.5} & {85.9} & {72.8} \\
{Mean pairwise overlap [\%]}        & {59.8} & {41.3} & {34.9} & {37.7} \\
{Resolution / FPS}             & \multicolumn{4}{c}{{\(640\times480\) RGB-D-semantic, \(30\,\)Hz}} \\
{Intrinsics (all agents)}      & \multicolumn{4}{c}{{\(f_x\!=\!f_y\!=\!320\), \(c_x\!=\!320\), \(c_y\!=\!240\) (\(90^{\circ}\) HFOV)}} \\

\bottomrule
\end{tabular}
\vspace{-1.2\baselineskip}
\end{table*}

{\subsection{Benchmark Statistics}
\label{sec:supp_stats}
Table~\ref{tab:benchmark_stats} summarizes the workload of each four-agent scene in ReplicaMultiagent Plus, including the number of input frames, travelled path length, and inter-agent overlap measured from the ground-truth depth by back-projecting and voxelizing the observations of all agents. Office-2 exhibits the highest mean pairwise overlap (\(59.8\%\)), as its agents share large portions of a compact office environment. In the larger Apartment-2 scene, the mean pairwise overlap decreases to \(34.9\%\), while \(85.9\%\) of the observed voxels remain visible to at least two agents. This indicates that most of the environment is jointly observed, but the overlap between any individual pair of agents is only partial, requiring global alignment to be assembled from multiple cross-agent overlaps. Hotel-0 provides an additional four-agent scene with a different spatial layout and overlap pattern, further broadening the diversity of the released benchmark. The quantitative baseline comparisons in Tables~\ref{tab:rendering_replica_plus} and~\ref{tab:geometry} use the three evaluation
scenes, Office-2, Apartment-0, and Apartment-2, on which all compared methods were evaluated. Hotel-0 is released as an additional four-agent scene and is fully characterized in Table~\ref{tab:benchmark_stats} and visualized in Fig.~\ref{fig:replica_plus_dataset}.}

{\section{Implementation Parameters}
\label{sec:supp_params}

Table~\ref{tab:params} summarizes the implementation parameters used throughout all experiments. For clarity, the parameters are organized according to the main stages of MAGS-SLAM: local monocular tracking and Gaussian mapping, keyframe and submap management, compact submap encoding and inter-agent retrieval, correspondence matching and geometric verification, global $\mathrm{Sim}(3)$ pose-graph optimization and MaPGBA consistency, occupancy-aware Gaussian fusion, and final global refinement. Parameters corresponding to quantities defined in the main text retain the same notation. They include the local mapping loss weights, keyframe and loop-selection thresholds, submap construction criteria, retrieval and matching gates, RANSAC/ICP verification thresholds, pose-graph weights and fusion eligibility criteria, the rigidity threshold used for reactive MaPGBA updates, and the voxelization and optimization settings used during fusion and refinement.}

\begin{table*}[t]
\caption{{Implementation parameters of MAGS-SLAM. Parameters are grouped by local mapping, submap construction, inter-agent retrieval and verification, global optimization, and Gaussian fusion.}}
\label{tab:params}
\centering
\scriptsize
\setlength{\tabcolsep}{4pt}
{%
\begin{tabular}{@{}lr@{\hspace{16pt}}lr@{\hspace{16pt}}lr@{}}
\toprule
\textbf{Parameter} & \textbf{Value} & \textbf{Parameter} & \textbf{Value} & \textbf{Parameter} & \textbf{Value} \\
\midrule
\multicolumn{6}{@{}l}{\textit{Front-end keyframes and intra-agent loop detection (Sec.~\ref{sec:intra_agent})}} \\
new keyframe: flow \(>\) & \(2.4\,\)px & drop keyframe: flow / covis.\,\(<\) & \(4.0\,\)px / 0.1 & end-of-run infill: covis.\ \(>\) & 0.2 \\
loop candidate: flow \(<\) & \(22\,\)px & loop cand.: min.\ keyframe gap & 52 & loop cand.: rot.\ \(<\) / max.\ no. & \(120^{\circ}\) / 10 \\
\midrule
\multicolumn{6}{@{}l}{\textit{Submap construction (Sec.~\ref{sec:intra_agent})}} \\
\(K_{\max}\) keyframes before freezing & 8 & \(\tau_{\text{move}}\) travel before freezing & \(2.0\,\)m & keyframes summarised & 3 \\
\(N_{\mathcal{Q}}\) salient points / summary & 512 & salient-point stride / min.\ dist. & 4 / 8\,px & \(\mathcal{R}^a_\ell\) cloud stride / pts / voxel & 2 / 4096 / \(5\,\)cm \\
\(\lambda_d\) / \(\lambda_F\) saliency weights & 0.35 / 0.10 \\
\midrule
\multicolumn{6}{@{}l}{\textit{Retrieval and correspondence (Sec.~\ref{sec:inter_agent})}} \\
\(K\) retrieval shortlist (top-\(K\)) & 3 & \(\tau_{\text{sim}}\) descriptor cosine gate & 0.30 & \(\bm{\theta}_{\text{match}}\) cosine / margin / Lowe & 0.55 / 0.02 / 0.90 \\
\(N_{\min}\) min.\ matches / inliers & 20 / 12 &  &  &  &  \\
\midrule
\multicolumn{6}{@{}l}{\textit{\(\mathrm{Sim}(3)\) estimation and verification (Sec.~\ref{sec:sim3_estimation})}} \\
RANSAC iterations / sample & 256 / 4 & \(\tau_{\text{in}}\) RANSAC inlier threshold & \(0.15\,\)m & \(r_{\min}\) minimum inlier ratio & 0.25 \\
\([s_{\min},s_{\max}]\) scale band & \([0.33, 3.0]\) & \(\tau_{\text{ext}}\) extent ratio, Eq.~\eqref{eq:extent_gate} & 0.15 & min.\ extent / degeneracy & \(0.30\,\)m / 0.01 \\
\(\Phi\) minimum ICP fitness & 0.25 & \(\rho_{\text{icp}}\) maximum ICP RMSE & \(0.20\,\)m & ICP corresp.\ dist.\ / iters & \(0.20\,\)m / 64 \\
\midrule
\multicolumn{6}{@{}l}{\textit{Submap graph optimisation (Sec.~\ref{sec:mapgba})}} \\
\(w^t\)/\(w^v\) temporal / inter-agent & 5.0 / 1.0 & \(w^{\text{pho}}_e\) photometric gain, verified & 1.0 & anchor stride / min.\ valid px & 2 / 64 \\
\(\rho\) robust loss / max.\ LM iters & Huber / 200 & \(\tau_{\text{res}}\) residual gate for fusion & \(0.20\,\)m & \(\tau_{\text{rig}}\) rigidity, Eq.~\eqref{eq:rigidity_fit_main} & \(0.10\,\)m \\
graph re-optimised every & 1 submap & Huber threshold on \(\sqrt{w}\,\mathbf{r}\) & 1.0 \\
\midrule
\multicolumn{6}{@{}l}{\textit{Occupancy-aware fusion (Sec.~\ref{sec:gs_fusion})}} \\
\(v\) occupancy voxel size & \(0.10\,\)m & \(k\) Gaussian envelope, Eq.~\eqref{eq:envelope} & 2.0 & ray stride / max.\ length & 16\,px / \(8.0\,\)m \\
min.\ new Gaussians, borrow & 30 & max.\ conflict ratio, borrow & 0.30 & opacity pre-filter, dedup. & 0.10 \\
\midrule
\multicolumn{6}{@{}l}{\textit{Mapping and global refinement (Eqs.~\eqref{eq:map_loss},~\eqref{eq:refine_loss})}} \\
\(\alpha,\beta\) photometric / depth w. & 0.95 / 0.25 & \(\lambda_n,\lambda_s\) normal / isotropy (map) & \(10^{-2}\) / 10 & \(\lambda\) SSIM w., refinement & 0.20 \\
\(\lambda_n'\) normal w., refinement & \(10^{-1}\) & refinement iters / LR scale & 2000 / 0.5 & mapping window / prune & 10 / 0.005 \\
\bottomrule
\end{tabular}}
\vspace{-1.0\baselineskip}
\end{table*}


\bibliographystyle{IEEEtran}
\bibliography{References}

\end{document}